\documentclass[acmlarge]{acmart}

\usepackage{booktabs} 

\setcopyright{usgovmixed}

\usepackage{graphicx}
\usepackage{subfig}

\usepackage{amsmath,amssymb,amsfonts}

\DeclareMathOperator*{\argmin}{arg\,min}

\usepackage[ruled,vlined]{algorithm2e} 

\SetAlFnt{\small}
\SetAlCapFnt{\small}
\SetAlCapNameFnt{\small}
\SetAlCapHSkip{0pt}
\IncMargin{-\parindent}

\usepackage{bbm}

\title[PaloBoost]{PaloBoost: An Overfitting-robust TreeBoost with Out-of-Bag Sample Regularization Techniques}

\author{Yubin Park}
\email{yubin.park@gmail.com}
\author{Joyce C. Ho}
\affiliation{%
  \institution{Emory University}
  \streetaddress{Mathematics and Science Center, W414}
  \city{Atlanta}
  \state{GA}
  \postcode{30322}
  \country{USA}}
\email{joyce.c.ho@emory.edu}

\begin{document}

\begin{abstract}
Stochastic Gradient TreeBoost is often found in many winning solutions in public data science challenges.
Unfortunately, the best performance requires extensive parameter tuning and can be prone to overfitting.
We propose PaloBoost, a Stochastic Gradient TreeBoost model that uses novel regularization techniques to guard against overfitting and is robust to parameter settings.
PaloBoost uses the under-utilized out-of-bag samples to perform gradient-aware pruning and estimate adaptive learning rates.
Unlike other Stochastic Gradient TreeBoost models that use the out-of-bag samples to estimate test errors, PaloBoost treats the samples as a second batch of training samples to prune the trees and adjust the learning rates.
As a result, PaloBoost can dynamically adjust tree depths and learning rates to achieve faster learning at the start and slower learning as the algorithm converges.
We illustrate how these regularization techniques can be efficiently implemented and propose a new formula for calculating feature importance to reflect the node coverages and learning rates.
Extensive experimental results on seven datasets demonstrate that PaloBoost is robust to overfitting, is less sensitivity to the parameters, and can also effectively identify meaningful features.
\end{abstract}



\renewcommand{\shortauthors}{Y. Park and J. C. Ho}

\maketitle

\section{Introduction}

Stochastic Gradient TreeBoost (SGTB) is one of the most widely used off-the-shelf machine learning algorithms \cite{Bekkerman:2015, He:2014, Chen:2016ga, Friedman:2006}.
SGTB is a stage-wise additive model where the base trees are fitted to the subsampled gradients (or errors) at each stage \cite{Friedman:2001ue, Friedman:2002ex}. 
The randomness introduced by subsampling speeds up the computation time and mitigates overfitting.
Consequently, SGTB can provide robust performance over various classification, regression, and even ranking tasks \cite{Burges:2010, Li:2008}.
Many empirical results have demonstrated SGTB's ability to model complex and large data relatively fast and accurately.
Furthermore, the effectiveness and pervasiveness of SGBT can be found in many winning solutions in public data science challenges \cite{Chen:2016ga}.

While SGTB can generally provide reasonable performance with the default parameter settings, to achieve its best performance usually requires extensive parameter tuning.
The four parameters in SGTB are maximum depth of the tree, number of trees, learning rate, and the subsampling rate.
While the maximum depth, subsampling rate, and learning rate have general guidelines, each parameter is not independent of one another. 
Having deeper trees and more trees can reduce training errors more rapidly, while potentially increasing the chance of overfitting \cite{Friedman:2001ue, Ridgeway:2012wr}.
Lower learning rates can mitigate overfitting to the training data at each stage, but more trees are needed to reach a similar level of performance \cite{Ridgeway:2012wr}. 
Thus, the delicate balancing act between overfitting and best performance is more of an art than a science in practice.

Exhaustive parameter tuning can be a bottleneck for many time-sensitive or performance-critical practical applications.
An ideal boosting algorithm should automatically learn what the four parameter values should be to achieve its best performance without overfitting.
Moreover, a smaller number of trees is equivalent to faster convergence, reduced probability of overfitting, and better performance.
This suggests two improvements: (1) the learning rate should be higher when the model is further away from convergence, and lower when the model is approaching its best performance and (2) the depth of the tree should change dynamically to account for the complexity of the target.
Thus the question is whether such a boosting algorithm can be built.

We present PaloBoost, a boosting algorithm that makes it easier to tune the parameters and potentially achieve better performance.
PaloBoost extends SGTB using two \textit{out-of-bag sample} regularization techniques: 1) Gradient-aware Pruning and 2) Adaptive Learning Rate.
Out-of-bag (OOB) samples, the samples not included from the subsampling process, are commonly available in many subsampling-based algorithms \cite{Breiman:1996, Breiman:2001}.
In boosting algorithms, OOB errors, or the errors estimated from OOB samples, are used as computationally cheaper alternatives for cross-validation errors and to determine the number of trees, also known as early stopping \cite{Ridgeway:2012wr}.
However, OOB samples are under-utilized in SGTB.
They merely play an observer role in the overall training process, often ignored and left unused.

PaloBoost, on the other hand, treats OOB samples as a second batch of training samples for adjusting learning rate and max tree depth.
At each stage of PaloBoost, the OOB errors are used to determine the generalization properties of the tree.
If the OOB does not decrease, the tree is too specific and likely overfit on the training data.
To mitigate the overfitting effect, the tree leaves are pruned to reduce the tree complexity, and optimal learning rates are estimated to control and decrease the OOB errors.
Thus, at each stage, PaloBoost randomly partitions the training samples into two sets: one is used for learning the tree structure, and the other is used to prune and adjust the learning rates.
This two-stage training process frequently appears in many machine learning algorithms to estimate two distinct sets of parameters or models \cite{Zhang:2018, Romero:2015, Barshan:2015}.
For example, in hyperparameter tuning \cite{Efron:1983, Kohavi:1995}, the parameters are sequentially estimated from two disjoint datasets:
(1) regular parameters from a training set and (2) hyperparameters from a validation set.
In PaloBoost, this translates to (1) in-bag samples for growing trees and (2) OOB samples for pruning and adjusting learning rates.
The main difference lies in the fact that PaloBoost's hyperparameter tuning occurs at each stage with different OOB samples.

We compared the performance of PaloBoost with various open-source SGTB implementations including Scikit-Learn \cite{scikit-learn} and XGBoost \cite{Chen:2016ga}.
Our benchmark datasets include one simulated data \cite{Friedman:1991} and six real datasets from Kaggle \cite{kaggle} and UCI repositories \cite{Dua:2017}.
The empirical results demonstrate stable, predictive performance in the presence of noisy features.
PaloBoost is considerably less sensitive to the hyperparameters and hardly suffers significant performance degradations with more trees.
The results also illustrate the adaptive learning rates and tree depths that guard against overfitting.
Moreover, we demonstrate the potential of our proposed feature importance formula to provide better feature selection than other SGTB implementations.
Thus, PaloBoost offers a robust, SGTB model that can save computation resources and researcher's time, and remove the art from hyperparameter tuning.
\section{Background}

Boosting algorithms build models in a stage-wise fashion, where the sequential learning of base learners provides a strong final model.
Breiman demonstrated that boosting can be interpreted as a gradient descent algorithm at the function level \cite{Breiman:1998ev}.
In other words, boosting is an iterative algorithm that tries to find the optimal ``function'', where the function is additively updated by fitting to the gradients (or errors) at each stage.
Later, Friedman formalized this view and introduced Gradient Boosting Machine (GBM) that generalizes to a broad range of loss functions \cite{Friedman:2001ue}.
Stochastic Gradient TreeBoost (SGTB) further builds on GBM to provide better predictive performance.
In this section, we provide the formulation and basics of GBM and illustrate how SGTB is derived.

\subsection{Gradient Boosting Machine}

GBM  \cite{Friedman:2001ue} seeks to estimate a function, $F^*$, that minimizes the empirical risk associated with a loss function $L(\cdot, \cdot)$ over $N$ pairs of target, $y$, and input features, $\mathbf{x}$:
\begin{equation}
F^* = \argmin_F \sum_{i=1}^N L(y_i, F(\mathbf{x}_i)) \label{eq:erm} .
\end{equation}
Examples of frequently used loss functions are squared error, $L(y,F) = (y-F)^2$ and negative binomial log-likelihood, $L(y, F) = -yF + \log(1+e^F)$, where $y \in \{0, 1\}$. 
GBM solves Equation \ref{eq:erm} by applying the gradient descent algorithm directly to the function $F$.
Each iterative update of the function $F$ has the form:
\begin{equation}
F_m(\mathbf{x}) = F_{m-1}(\mathbf{x}) + \beta_m \frac{\partial}{\partial F(\mathbf{x})} L(y, F(\mathbf{x})) \Bigr|_{F=F_{m-1}} \label{eq:gbmupdateideal},
\end{equation}
where $\beta_m$ is the step size.
With finite samples, we can only approximate the gradient term with an approximation function, $h(\mathbf{x})$:
\begin{equation}
F_m(\mathbf{x}) = F_{m-1}(\mathbf{x}) + \beta_m h(\mathbf{x}; \mathbf{a}_m) \label{eq:gbmiter},
\end{equation}
where $\mathbf{a}_m$ represents the parameters for the approximation function at iteration $m$.
In GBM, $h(\mathbf{x}; \mathbf{a})$ can be any parametric function such as neural net \cite{Schwenk:2000}, support vector machine \cite{Ting:2009}, and regression tree \cite{Friedman:2001ue}.
Algorithm~\ref{algo:gbm} outlines the details of GBM.

\begin{algorithm}
\DontPrintSemicolon
$F_0 = \argmin_{\beta} \sum_{i=1}^N L(y_i, \beta)$\;
\For{$m \gets 1$ \textbf{to} $M$}{
    $z_i = -\frac{\partial}{\partial F(\mathbf{x}_i)} L(y_i, F(\mathbf{x}_i)) \Bigr|_{F=F_{m-1}}$, for $i=1, \cdots, N$\;
    $\textbf{a}_m = \argmin_\textbf{a} \sum_{i=1}^N \| z_i - h(\mathbf{x}_i; \mathbf{a})\|^2$\;
    $\beta_m = \argmin_\beta \sum_{i=1}^N L(y_i, F_{m-1}(\mathbf{x}_i) + \beta h(\mathbf{x}_i; \mathbf{a}_m))$ \;
    $F_m(\mathbf{x}) = F_{m-1}(\mathbf{x}) + \beta_m h(\mathbf{x}; \mathbf{a}_m)$\;
}
\caption{Gradient Boosting Machine with Base Learner $h(\mathbf{x}; \mathbf{a})$}
\label{algo:gbm}
\end{algorithm}

An alternative perspective of GBM is to view it as an ensemble where the approximation function, $h(\mathbf{x})$, is the base learner and $\beta_m$ represents the corresponding weight.
The GBM model (outlined in Algorithm~\ref{algo:gbm}) produces a final output $F$ of the form:
\begin{equation}
F(\mathbf{x}) = F_0 + \sum_{m=1}^M \beta_m h_m(\mathbf{x}) \label{eq:addm}.
\end{equation}
In fact, XGBoost, a variant of GBM, uses this approach to develop a scalable and flexible boosting system \cite{Chen:2016ga}.
By assuming that $F$ takes the ensemble form, the optimal base learners are derived by applying the Taylor approximation and a greedy optimization technique.
PaloBoost is better understood in the context of the original GBM work \cite{Friedman:2001ue}.
Thus, the notations and formulation will follow Friedman's work.

\subsection{Stochastic Gradient TreeBoost}

Stochastic Gradient TreeBoost (SGTB) introduces two important modifications to GBM: (1) tree structure-aware step sizes and (2) subsampling at each stage \cite{Friedman:2001ue}.
Friedman observed that a tree partitions the input into $J$ disjoint regions ($\{R_j\}_1^J$), where each region predicts a constant value ($b_j$).
Thus, the approximation function can be expressed as:
\begin{equation}
h(\mathbf{x}; \mathbf{a}) = h(\mathbf{x}; \{b_j, R_j\}_1^J) = \sum_{j=1}^J b_j \mathbbm{1}(\mathbf{x} \in R_j).
\end{equation}
This representation allows each region to choose its own optimal step size, $\beta_{jm}$, instead of a single step size per stage $\beta_m$ to obtain a better predictive model.
Therefore, Equation \ref{eq:gbmiter} can be rewritten as:
\begin{equation}
F_m(\mathbf{x}) = F_{m-1}(\mathbf{x}) +  \sum_{j=1}^J \gamma_{jm} \mathbbm{1}(\mathbf{x} \in R_{jm})
\end{equation}
where $\gamma_{jm} = \beta_{jm} b_j $.
As a result, estimating $\gamma_{jm}$ is equivalent to estimating the intercept that minimizes the loss function within the disjoint region $R_{jm}$:
\begin{equation}
\gamma_{jm} = \argmin_{\gamma} \sum_{\mathbf{x}_i \in R_{jm}} L(y_i, F_{m-1}(\mathbf{x}_i) + \gamma).
\end{equation}

The second important modification in SGTB is subsampling, which is motivated by bagging \cite{Breiman:1996}  and AdaBoost \cite{Freund:1999}.
Friedman found that random subsampling of the training samples greatly improved the performance and generalization abilities of GBM, while reducing the computing time.
Although the performance improvement varied across problems, he observed that subsampling has a greater impact on small datasets with high capacity base learners.
Consequently, he postulated that the performance improvement may be due to the variance reduction as in bagging.
Typical subsampling rates ($q$) lie between 0.5 and 0.7.

However, subsampling alone may be insufficient.
A shrinkage technique was introduced to scale the contribution of each tree by a factor $\nu$ between $0$ and $1$.
This parameter can also be viewed as the learning rate $\nu$ in the context of stochastic gradient descent.
Lower $\nu$ values ($\le 0.1$) slow down the learning speed of SGTB and would need more iterations, 
but may achieve better generalization \cite{Friedman:2002ex}.
Algorithm~\ref{algo:treeboost} illustrates the details of SGTB with a learning rate $\nu$.
Thus, SGTB has four different parameters that require tuning: (1) tree size (related to tree depth) ($J$); (2) number of trees ($M$); (3) the learning rate ($\nu$); and (4) the subsampling rate ($q$).  

\begin{algorithm}
\DontPrintSemicolon
$F_0 = \argmin_{\beta} \sum_{i=1}^N L(y_i, \beta)$\;
\For{$m \gets 1$ \textbf{to} $M$}{
    $\{y_i, \mathbf{x}_i\}_1^{N^\prime} = \text{Subsample}(\{y_i, \mathbf{x}_i\}_1^{N}, \text{rate}=q)$\;
    $z_i = -\frac{\partial}{\partial F(\mathbf{x}_i)} L(y_i, F(\mathbf{x}_i)) \Bigr|_{F=F_{m-1}}$, for $i=1, \cdots, N^\prime$\;
    $\{R_{jm}\}_1^J = \text{RegressionTree}(\{z_i, \mathbf{x}_i\}_i^{N^\prime})$\;
    $\gamma_{jm} = \argmin_{\gamma} \sum_{\mathbf{x}_i \in R_{jm}} L(y_i, F_{m-1}(\mathbf{x}_i) + \gamma)$, for $j=1, \cdots, J$ \;
    $F_m(\mathbf{x}) =F_{m-1}(\mathbf{x}) +  \nu \sum_{j=1}^J \gamma_{jm} \mathbbm{1}(\mathbf{x} \in R_{jm})$\;
}
\caption{Stochastic Treeboost with Learning Rate ($\nu$)}
\label{algo:treeboost}
\end{algorithm}

SGTB is widely available in many easy-to-use open-source packages.
Some implementations include Generalized Boosted Models available in \texttt{R} \cite{Ridgeway:2012wr}, TreeNet by Salford Systems \cite{treenet}, Gradient Boosting in Scikit-Learn \cite{scikit-learn}, and XGBoost implemented in \texttt{C++} \cite{Chen:2016ga}.
Although all packages follow Algorithm~\ref{algo:treeboost}, the implementations are slightly different.
As an example, the splitting criteria of the base learner varies across packages -- Generalized Boosted Models uses the minimum variance criterion, Scikit-Learn uses the Friedman splitting criterion \cite{Friedman:2001ue} by default, and XGBoost uses a custom regularized splitting criterion that is obtained by greedily minimizing the Taylor approximation of the loss function \cite{Chen:2016ga}.
The treatment of missing values can also differ (e.g., XGBoost can handle missing values by default, while Scikit-Learn does not support it).
Thus, the performance can vary across packages even when applied to the same dataset.
However, there is no conclusive evidence that one implementation is the best, as the performance differences are dataset-dependent.

\section{PaloBoost}\label{sec:paloboost}

The performance of SGTB is dependent on its hyperparameters: tree depth, number of trees, subsampling rate, and learning rate.
Unfortunately, each parameter is not independent of the others and makes the tuning process quite difficult.
As an example, the $\nu$-$M$ trade-off \cite{Friedman:2001ue, Friedman:2002ex}, captures the relationship between the learning rate and the number of trees.
Lower learning rates ($\nu$) may result in better performance, but more trees ($M$) are usually required to achieve a similar level of performance. 
Thus, an exhaustive grid search is often needed to find the optimal hyperparameters.

Although the hyperparameter tuning process is relatively straightforward, it can consume a substantial amount of computational resources.
We illustrate the tuning process for tree depth and learning rate for a \emph{single-stage} SGTB model (i.e., $M = 1$).
The dataset is partitioned into three sets: training, validation, and test.
The test set is strictly set aside and used to estimate the generalization error.
First, the intercept term, $F_0$, is fit using the training set.
Next, for a given learning rate and tree depth, we randomly subsample data points from the training set and fit a tree on those sampled gradients.
The performance is then measured on the validation set.
This process is repeated multiple times using varying learning rates and tree depths.
The hyperparameters are then chosen based on the best validation performance.
The sample principle is applied for SGTBs with multiple stages, where the number of trees ($M$) is also varied.
As a result, the whole cycle, while parallelizable, is computationally expensive and requires considerable time.

Unfortunately, even the best-tuned hyperparameters may not produce optimal results in real testing environments.
Since the training and validation samples may not reflect the ``true" distribution, there is a real danger of overfitting.
Thus, the hyperparameters chosen via the expensive tuning process may not yield a robust, and generalizable model.
We have developed PaloBoost, a variant of SGTB that is less sensitive to hyperparameters and provides robust predictive performance. 
PaloBoost eliminates the need to finely tune the learning rate and tree depth, and instead estimates them during the training process.
Moreover, our implementation provides comparable training times to existing state-of-the-art SGTB implementations.

\begin{figure}
    \centering
        \includegraphics[width=1.0\columnwidth]{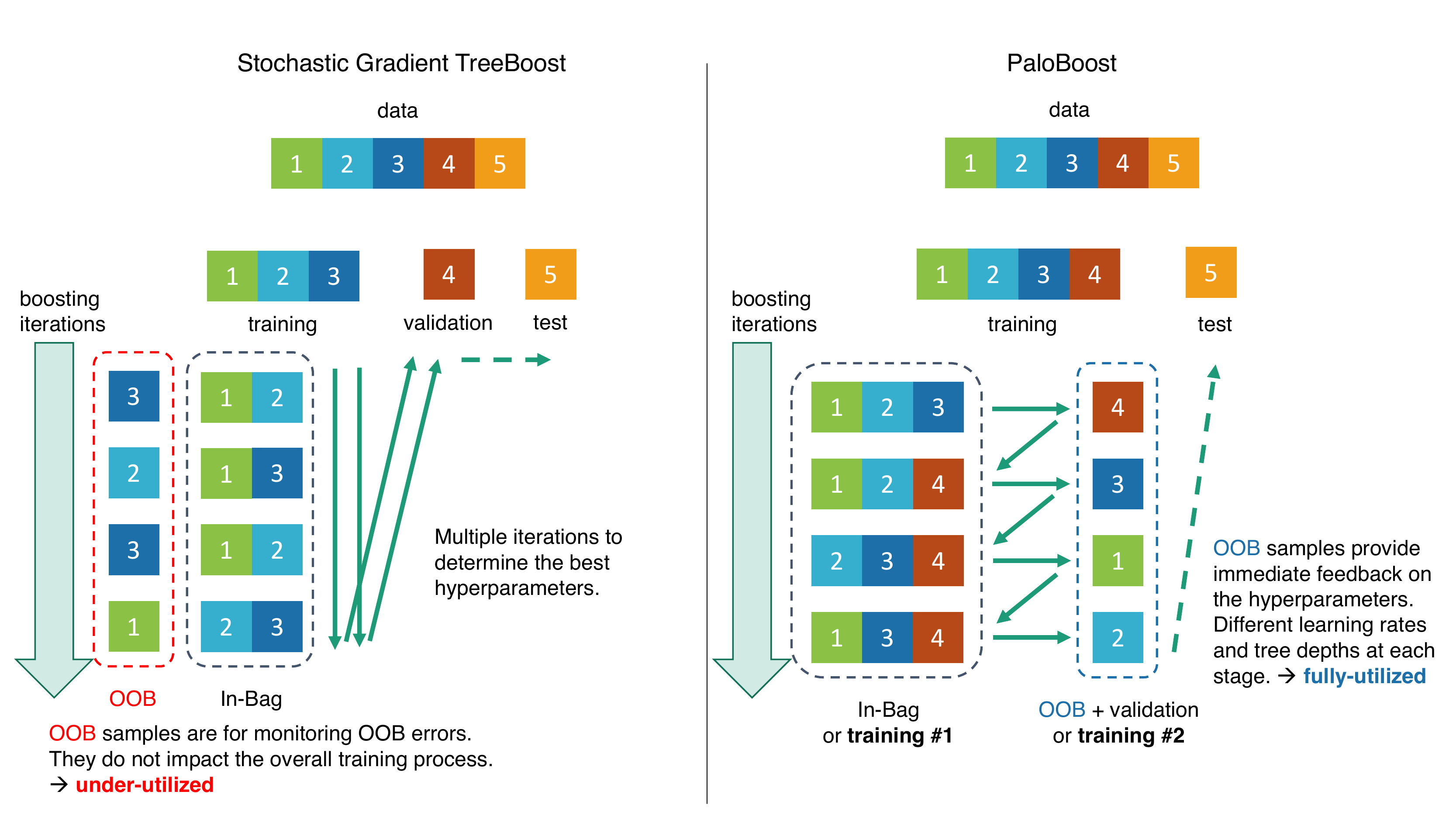}
        \caption{Data flows in SGTB and PaloBoost.}
    \label{fig:dataflow}
\end{figure}

PaloBoost mitigates overfitting via adaptive learning rates and tree depths.
The main idea centers around the \textit{under-utilized} out-of-bag (OOB) samples to tune these parameters.
We observed that \textit{both validation and OOB samples} are not seen during the training phase.
Thus, the OOB samples can be used to estimate the learning rate and the tree depth at each individual stage.
The sample principle can be applied at each stage even though the OOB samples are different as the learning rate and tree depth are tree-specific parameters.
Additionally, the stage-specific adaptive learning rates serve as a guard against overfitting and can be used to determine the optimal number of trees.
Thus, PaloBoost does not need to maintain a separate validation set, thereby increasing the number of training samples at each stage that can further combat tree overfitting.
Figure \ref{fig:dataflow} illustrates the conceptual differences in the learning process for multiple-stages SGTB and PaloBoost.

\subsection{Gradient-Aware Pruning}

The maximum depth of the trees is a hyperparameter for many SGTB implementations.
While the tree at each stage may not grow to the maximum depth for various reasons (e.g., insufficient samples in the node, information gain is not above a tolerance, etc.), the ``intention" is to maintain the maximum height.
Consequently, the majority of the trees in SGTB implementations will be grown to the maximum tree depth and can result in overfitting.
Although XGBoost has a pre-pruning regularization parameter \cite{Chen:2016ga}, our empirical studies showed little impact unless the default parameter was adjusted substantially.
Instead, we introduce a gradient-aware pruning technique that utilizes the OOB samples to achieve more flexible tree depths.
Even though maximum depth remains a parameter in PaloBoost, our model is less sensitive to the value and does not need to be finely tuned.

Gradient-aware pruning has roots in the bottom-up ``reduced-error pruning`` found in decision tree literature.
In reduced-error pruning, the errors on OOB samples are compared between children and parent nodes.
If the child node does not decrease the error, the node is pruned, thereby reducing the complexity of the tree.
However, boosting poses two challenges: each tree is dependent on the previous trees, and the node estimates are always multiplied by the learning rate.
Therefore, a new pruning approach is necessary to account for these aspects.

Conceptually, gradient-aware pruning removes gradient estimates that do not generalize well to other samples.
As the tree depth increases, the number of samples per node decreases.
Smaller samples result in higher variances of the estimated gradient, $\gamma$, which is also likely to increase generalization errors.
Thus, to achieve more stable gradient estimates, the regions with high variance should be merged.
In PaloBoost, after a tree is fitted to the subsampled gradients, the tree is applied to the OOB samples.
For each disjoint region of the tree $(R_j)$, the loss associated with introducing a new leaf estimate is the gradient multiplied by the learning rate:
\begin{equation}
\text{Loss}(R_j) = \sum_{i}^{L_j} \text{Loss}(y_i, \hat{y}_i + \nu \gamma_j)
\label{eq:prune}
\end{equation}
Thus, if the leaf estimate does not reduce the loss on the OOB samples, the node should be pruned.
The gradient-aware pruning process is summarized in Algorithm~\ref{algo:prune}.

\begin{algorithm}
\DontPrintSemicolon
$\{(R_j, R_k)\} = \text{Find-Sibling-Pairs}(\{R_j\}_{j=1}^J)$\;
\For{each sibling pair in $\{(R_j, R_k)\}$}{
    $\{y_i, \mathbf{x}_i\}_1^{L_j} = \text{Out-of-Bag}(\{y_i, \mathbf{x}_i\}_1^{N} | R_j)$\;
    $\{y_i, \mathbf{x}_i\}_1^{L_k} = \text{Out-of-Bag}(\{y_i, \mathbf{x}_i\}_1^{N} | R_k)$\;    
    \uIf{$\sum_{i}^{L_j} \text{Loss}(y_i, \hat{y}_i) < \sum_{i}^{L_j} \text{Loss}(y_i, \hat{y}_i + \nu_{max} \gamma_j)$}{
        do\_merge = True\;        
    }\uElseIf{$\sum_{i}^{L_k} \text{Loss}(y_i, \hat{y}_i) < \sum_{i}^{L_k} \text{Loss}(y_i, \hat{y}_i + \nu_{max} \gamma_k)$}{
        do\_merge = True\;            
    }\Else{ 
        do\_merge = False\; 
    }
    \If{do\_merge}{
        Merge($R_j$, $R_k$)
    }
}
\caption{Gradient-Aware Pruning}
\label{algo:prune}
\end{algorithm}

\subsection{Adaptive Learning Rate}

The learning rate, $\nu$, controls the contribution of each new stage.
Empirically, the best strategy is to set $\nu$ to a very small value to achieve favorable test errors at the cost of larger values of $M$ \cite{Hastie:esl}.
However, a single learning rate for every tree may not be optimal, as some trees do not generalize well.
Instead, each stage should have a different learning rate.
Unfortunately, calculating optimal stage-specific learning rates through standard optimization techniques (i.e., line search) introduces substantial computational overhead.
Rather, PaloBoost takes advantage of the tree structure to calculate optimal learning rates for each region efficiently.

The key observation is that we can decouple the estimation of the learning rate if we introduce a region-specific learning rate at each stage.
Thus, each region $R_{j}$ can calculate the multiplicative factor that optimizes the OOB loss:
\begin{equation} 
\nu_j^* = \argmin_\nu \sum_{i}^{L_j} Loss(y_i, \hat{y}_i +  \nu\gamma_j),
\end{equation}
where $L_j$ represents the number of OOB samples in the leaf of interest.
The benefit of this approach is that closed form solutions exist for many loss functions.
For example, the learning rate for the negative binomial log-likelihood loss function is:
\begin{align}
\nu_j^*&= \argmin_\nu \sum_{i}^{L_j} \log(1+\exp(\hat{y}_i + \nu\gamma_j)) - y_i (\hat{y}_i + \nu\gamma_j) \\
&= \log \left(\frac{\sum_{i}^{L_j} y_i }{ \sum_{i}^{L_j} (1-y_i)\text{exp}(\hat{y}_i)} \right)/\gamma_j
\end{align}
Similarly, for the squared error loss function, the closed form solution is:
\begin{align}
\nu_j^* &= \argmin_\nu \sum_{i}^{L_j} (y_i - (\hat{y}_i + \nu\gamma_j))^2 \\
&= \frac{\sum_{i=1}^{L_j}(y_i - \hat{y}_i) }{ \gamma_j L_j}
\end{align}

PaloBoost also introduces a clipping function on the estimated learning rate, to reduce the effect of small OOB sample sizes.
Without clipping, the estimated learning rates can fluctuate in a wide range, sometimes exceeding the value 1.
Thus, to enforce stability and maintain similarity with existing SGTB implementations, estimated learning rates are capped with the maximum learning rate parameter $\nu_{max}$.
As a result, the effective region-specific learning rate ranges between zero and the maximum specified learning rate.
Algorithm~\ref{algo:alr} illustrates our adaptive learning rate strategy. 

\begin{algorithm}
\DontPrintSemicolon
$\{(R_j, R_k)\} = \text{Find-Sibling-Pairs}(\{R_j\}_{j=1}^J)$\;
\For{$R_j$ in $\{R_j\}_{j=1}^J$}{
    $\{y_i, \mathbf{x}_i\}_1^{L_j} = \text{Out-of-Bag}(\{y_i, \mathbf{x}_i\}_1^{N} | R_j)$\;
    \uIf{distribution==``gaussian``}{
        $\nu_j =  \text{clip}\left(\frac{\sum_{i=1}^{L_j}(y_i - \hat{y}_i) }{ \gamma_j L_j}, 0, \nu_{max} \right)$\;        
    }\ElseIf{distribution==``bernoulli``}{
        $\nu_j = \text{clip}\left( \log \left(\frac{\sum_{i}^{L_j} y_i }{ \sum_{i}^{L_j} (1-y_i)\exp(\hat{y}_i)}\right)/\gamma_j , 0, \nu_{max} \right)$\;    
    }
}
\caption{Adaptive Learning Rate with Out-of-Bag Loss Reduction}
\label{algo:alr}
\end{algorithm}

Gradient-aware pruning serves as a preprocessing step for the adaptive learning rate mechanism.
Removing the regions with higher variances of node estimates ($\gamma$) provides two main benefits: (1) there are less adaptive rates to estimate, and (2) robust node estimates will yield better learning rates.
Algorithm \ref{algo:paloboost} provides the details of PaloBoost.
As learning rate and tree depth values are re-estimated during the training process, there is less sensitivity to the maximum learning rate parameter $\nu_{max}$ and tree depth.
Moreover, these modifications also provide a safeguard for a variety of subsampling rates.
Therefore, only the number of trees, $M$, requires ``extensive'' tuning.

\begin{algorithm}
\DontPrintSemicolon
$F_0 = \argmin_{\beta} \sum_{i=1}^N L(y_i, \beta)$\;
\For{$m \gets 1$ \textbf{to} $M$}{
    $\{y_i, \mathbf{x}_i\}_1^{N^\prime} = \text{Subsample}(\{y_i, \mathbf{x}_i\}_1^{N}, \text{rate}=q)$\;
    $z_i = -\frac{\partial}{\partial F(\mathbf{x}_i)} L(y_i, F(\mathbf{x}_i)) \Bigr|_{F=F_{m-1}}$, for $i=1, \cdots, N^\prime$\;
    $\{R_{jm}\}_1^J = \text{RegressionTree}(\{z_i, \mathbf{x}_i\}_i^{N^\prime})$\;
    $\gamma_{jm} = \argmin_{\gamma} \sum_{\mathbf{x}_i \in R_{jm}} L(y_i, F_{m-1}(\mathbf{x}_i) + \gamma)$, for $j=1, \cdots, J$ \;
    $\{R_{jm}, \gamma_{jm}\}_1^{J^\prime} = \text{Gradient-Aware-Pruning}(\{R_{jm}, \gamma_{jm}\}_1^J, \nu_{max})$\;
    $\nu_{jm} = \text{Learning-Rate-Adjustment}(\gamma_{jm}, \nu_{max})$, for $j=1, \cdots, J^\prime$\;
    $F_m(\mathbf{x}) =F_{m-1}(\mathbf{x}) +   \sum_{j=1}^J \nu_{jm} \gamma_{jm} \mathbbm{1}(\mathbf{x} \in R_{jm})$\;
}
\caption{PaloBoost}
\label{algo:paloboost}
\end{algorithm}

\subsection{Modified Feature Importance}\label{sec:paloboost-feaures}

Feature importance in SGTB is calculated as the summation of squared error improvements by the feature \cite{Elith:2008}.
Although the formula is adopted from RandomForest \cite{Breiman:2001}, the measure is purely based on heuristic arguments \cite{Friedman:2001ue}.
While the feature importance has proven useful in many applications, we notice that ignores two important aspects of SGTB: the leaf estimates and the coverage of the region.
As seen in the SGTB algorithm (Algorithm \ref{algo:treeboost}), leaf estimates are recalibrated to minimize the loss function.
Moreover, the improvement from a feature at a particular branch may yield leaves that have extremely small coverage. 
Thus, the feature importance may not be an accurate indication of coverage and impact.

These issues, along with the new adaptive learning rate, led us to create a new feature importance formula.
Instead of a branch-centric perspective, PaloBoost introduces a leaf-centric formula that accounts for leaf estimate, region coverage, and region-specific learning rate.
The importance of a feature, $f_k$ is calculated as:
\begin{equation}
f_k = \sum_{m=1}^{M}  \frac{ \sum_{j=1}^{J_m} \nu_{jm} | R_{jm} |  | \gamma_{jm} |  \mathbbm{1}(f_k \in \text{Rules}_{jm})}{J_m \sum_{j=1}^{J} R_{jm}}
\end{equation}
where $|R_{jm}|$ represent the coverage of the region $R_{jm}$, 
and $\text{Rules}_{jm}$ denotes the logical rules that define the region $R_{jm}$.
For example, if $\text{Rules}_{jm}$ is of the form $(x_0 > 0.5 \text{ AND } x_1 < 1.0 \text{ AND } x_3 > -1)$, then $\mathbbm{1}(f_1 \in \text{Rules}_{jm})$ will be one as $x_1$ is in the rules.
With our feature importance formula, if a feature is used to define a region, we will multiply the size of the region ($|R_{jm}|$) with the effective absolute node estimate, $\nu_{jm}| \gamma_{jm}|$. 
Therefore, if any of the three quantities (coverage of the region, the leaf estimate, or the adaptive learning rate) is very small, the feature importance contribution is minimal.

\section{Experimental Results}

PaloBoost is implemented in Python using the Bonsai Decision Tree framework \cite{bonsai-dt}, which allows easy manipulation of decision trees.
The Bonsai framework is mostly written in Python with its core computation modules written in Cython, to provide C-like performance.
Our implementation of PaloBoost is made publicly available as an extension to Bonsai\footnote{The implementation of PaloBoost can be found in \url{https://yubin-park.github.io/bonsai-dt/} as one of the Bonsai templates.}.

PaloBoost is evaluated on a variety of datasets (simulated and real-world) and compared against state-of-the-art implementations.
In this section, we will analyze the following aspects of PaloBoost:
\begin{itemize}
\item How different are the learning rates and tree depths at each individual stage?
\item How sensitive is PaloBoost to the specified hyperparameters?
\item How does the predictive performance compare with other SGTB implementations?
\item Is PaloBoost more computationally expensive than existing SGTB implementations?
\end{itemize}
All experiments are carried out on a single machine running macOS High Sierra 10.13.6 with 2.7 GHz Intel Core i5 CPU and 8 GB of RAM.

\subsection{Datasets}

Seven different datasets are used to evaluate the SGTB variants, including PaloBoost.
We use one simulated dataset and six publicly available datasets, two from the UC Irvine (UCI) machine learning repository \cite{Dua:2017} and four from Kaggle \cite{kaggle}, a popular data science competition platform.
Our benchmark study consists of three regression tasks and four classification tasks.
The sizes of the datasets range from a few thousand samples (1,994 samples) to a little bit over a hundred thousand samples (114,321 samples).
Small datasets are included to highlight the performance sensitivity with respect to the hyperparameters.
Table \ref{tab:data} lists the seven real-world datasets and their associated characteristics.
A brief overview of each dataset will be provided in the context of their respective tasks.

\begin{table}
\centering
\caption{Benchmark Datasets. For regression tasks, we list the standard deviation of the target variable, whereas the class ratios for classification tasks. The ``\# Real'' and ``\# Cat'' columns represent the number of numeric and categorical features, respectively. The ``\# Final'' column shows the final number of features after our pre-processing step.}
\label{tab:data}
\begin{tabular}{l  l   l   r   r   r   r   r} 
\toprule
Dataset & Task & Missing Data & Target Stats & \# Samples & \# Real & \# Cat & \# Final  \\
\midrule
Friedman-Sim \cite{Friedman:1991} & REG & None & $\sigma(y)=6.953$ & 10,000 & 10 & 0 & 10  \\
Mercedes-Benz \cite{kaggle:mercedes} & REG & None & $\sigma(y)=12.67$ & 4,209 & 368 & 8 & 460 \\
Crime-Rate \cite{Redmond:2002} & REG & Present & $\sigma(y)=0.233$ & 1,994 & 126 & 1 & 127 \\
Amazon-Empl \cite{kaggle:amazon} & CLS & None & $\mu(y)=0.942 $ & 32,769 & 0 & 9 & 115 \\
Pulsar-Detect \cite{Lyon:2016, Lyon:data} & CLS & None & $\mu(y)=0.092 $ & 17,898 & 8 & 0 & 8 \\
Carvana \cite{kaggle:carvana} & CLS & Present & $\mu(y)=0.123$ & 72,983 & 17 & 14 & 151 \\
BNP-Paribas \cite{kaggle:bnp} & CLS & Present & $\mu(y)=0.761$ & 114,321 & 112 & 19 & 273 \\
\bottomrule
\end{tabular}
\end{table}

Many of the datasets contain both numeric and categorical features.
For the categorical features, the top 20 frequent categories are selected and transformed using one-hot encoding (also known as dummy-coding).
To minimize the impact of preprocessing, no additional techniques were applied (e.g. imputation, outlier detection).
While XGBoost and our SGTB implementations (including PaloBoost) can naturally handle missing values, we omit Scikit-Learn for datasets with missing values as it does not support missing values.

\subsection{Baselines}

We compare PaloBoost with three other baseline models as follows:
\begin{itemize}
\item Scikit-Learn \cite{scikit-learn}, the de-facto machine learning library in Python that has been widely adopted.
\item XGBoost \cite{Chen:2016ga}, the battle-tested library that has won many data science challenges.
\item Bonsai-SGTB, an SGTB implementation using the Bonsai framework \footnote{\url{https://yubin-park.github.io/bonsai-dt/}}.
\end{itemize}
While there are many newly developed SGTB implementations (e.g., LightGBM, CatBoost), they are not included in our study for several reasons.
First, we observed that the overfitting behavior of these packages are similar to the Scikit-Learn and XGBoost as they focus primarily on implementation details and feature engineering.
CatBoost focuses on more efficiently encoding categorical variables using Target-based Statistics and permutation techniques \cite{Prokhorenkova:2017ul, Dorogush:2017vz}.
LightGBM primarily focuses on training speed using less memory in distributed settings \cite{Ke:2017ut}. 
Secondly, our objective is to characterize the behaviors of the two regularization techniques in PaloBoost, rather than to claim that PaloBoost outperforms other SGTB implementations.
We note that our regularization techniques, Gradient-aware Pruning and Adaptive Learning Rate, can be added to any existing SGTB implementations.
Therefore, the three baseline models serve as suitable representations for other implementations.

\subsection{Setup \& Evaluation Metrics}

All datasets are split with a 30/70 train-test ratio (i.e., 30\% for training and 70\% for test). 
While we performed extensive experiments using other ratio choices and found similar behaviors, this specific choice showcases the sensitivity of SGTB to hyperparameters.
With larger training ratios (e.g., 50/50, 60/40, 70/30, 80/20), we found that it was difficult to visualize the overfitting behaviors for many of the baseline models.
On the contrary, for smaller training ratios such as 10/90 and 20/80, the results were significantly more favorable to PaloBoost.
The 30/70 split best illustrated the characteristics of PaloBoost without significantly sacrificing the training sample size.

For each SGTB implementation, we tested three different learning rates that spanned the spectrum for $\nu$ (1.0, 0.5, and 0.1).
Also for consistency across the models, the tree depth was set to be five levels (note that PaloBoost will have lower depths due to Gradient-aware pruning).
Unless otherwise specified, default parameter settings were used for all our benchmark models.
The trained models (on the 30\% training set) are then applied to the test set where we track the predictive performance.
This is measured using the coefficient of determination ($R^2$) and the Area Under the Receiver Operating Characteristic Curve (AUROC) for the regression and classification tasks, respectively.
We also track two different metrics for PaloBoost:
\begin{itemize}
\item \textbf{Average Learning Rate}:
For visualization purposes, we defined the average learning rate per stage to compress the different learning rates for each disjoint region of a tree.
This is a weighted average of the learning rates, where the weights are proportional to the leaf coverage, and is defined as:
\begin{equation}
\bar{\nu}_m =\frac{\sum_j \nu_{jm} | R_{jm} | }{\sum_j | R_{jm} |}
\end{equation}
\item \textbf{Prune Rate}: We measure the number of nodes removed by the Gradient-aware pruning mechanism.
The prune rate is defined as follows:
\begin{equation}
\text{Prune Rate} = \frac{J - J^\prime}{J}
\end{equation}
where $J$ and $J^\prime$ represent the number of disjoint regions in a tree before and after Gradient-aware Pruning, respectively.
\end{itemize}

\subsection{Simulated Dataset}

\emph{Description}. The simulated dataset is adopted from Friedman \cite{Friedman:1983, Friedman:1991}, to introduce noisy features (i.e., features with zero importance).
The formula for the dataset is as follows:
\begin{align}
y &= 10 \sin(\pi  x_0  x_1) + 20 (x_2-0.5)^2 + 10 x_3 + 5 x_4 + 5\epsilon \label{eq:friedman}\\
\epsilon &\sim \text{Normal}(0, 1)\\
x_i &\sim \text{Uniform}(0,1) \text{ where } i = 0 \ldots 9 
\end{align}
where $x_i$s are uniformly distributed on the interval [0,1].
For each sample, Gaussian noise $\epsilon$ with a standard deviation of five, is added.
Note that only 5 features among 10 features ($x_0 - x_4$) are used to generate the target ($y$) in Equation~\ref{eq:friedman}.
Our simulated dataset contains 10,000 samples.

\begin{figure}[tb]
    \centering
        \subfloat[$R^2$, $\nu_{max}=1.0$]{\includegraphics[width=0.33\textwidth]{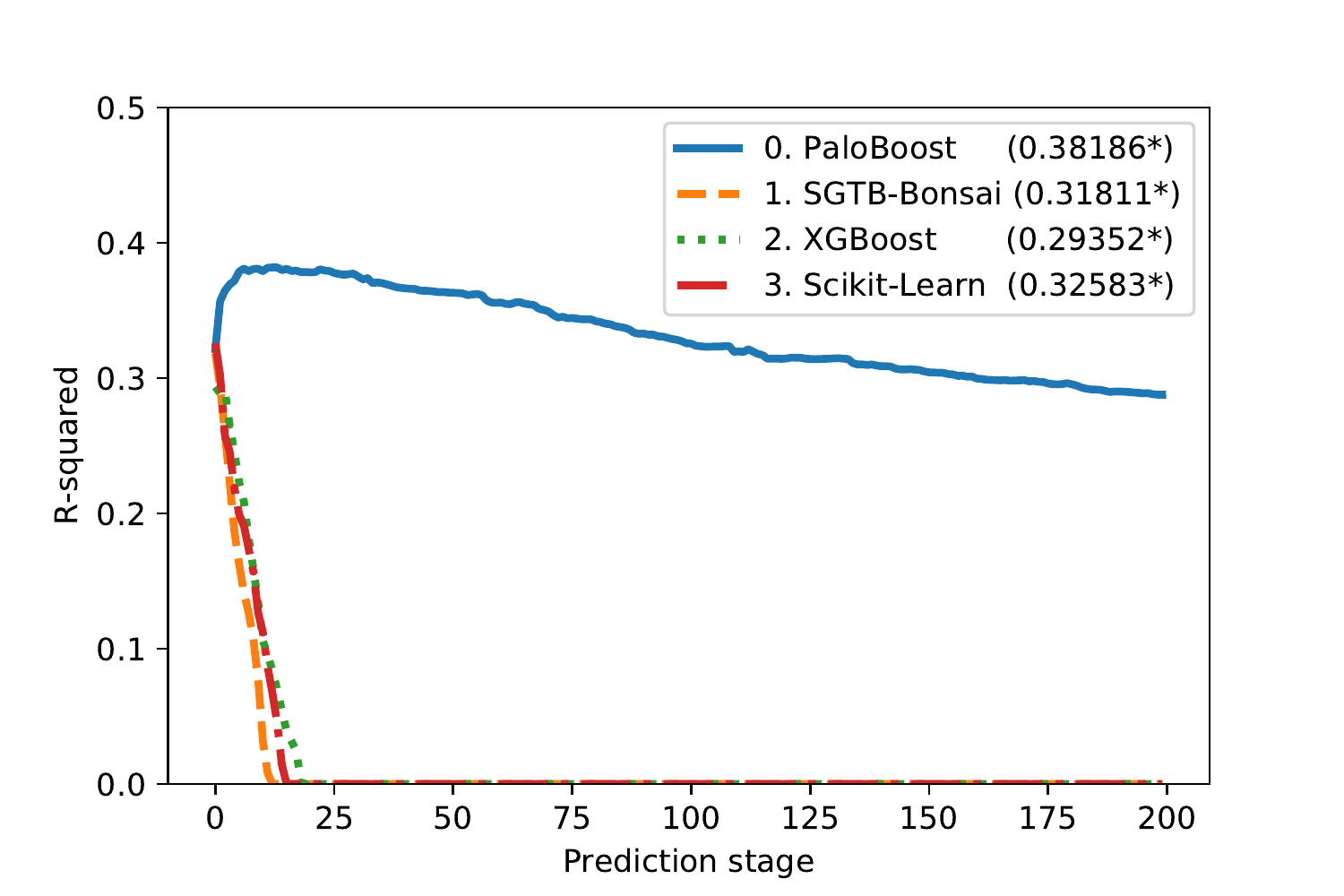}}
        \subfloat[$R^2$, $\nu_{max}=0.5$]{\includegraphics[width=0.33\textwidth]{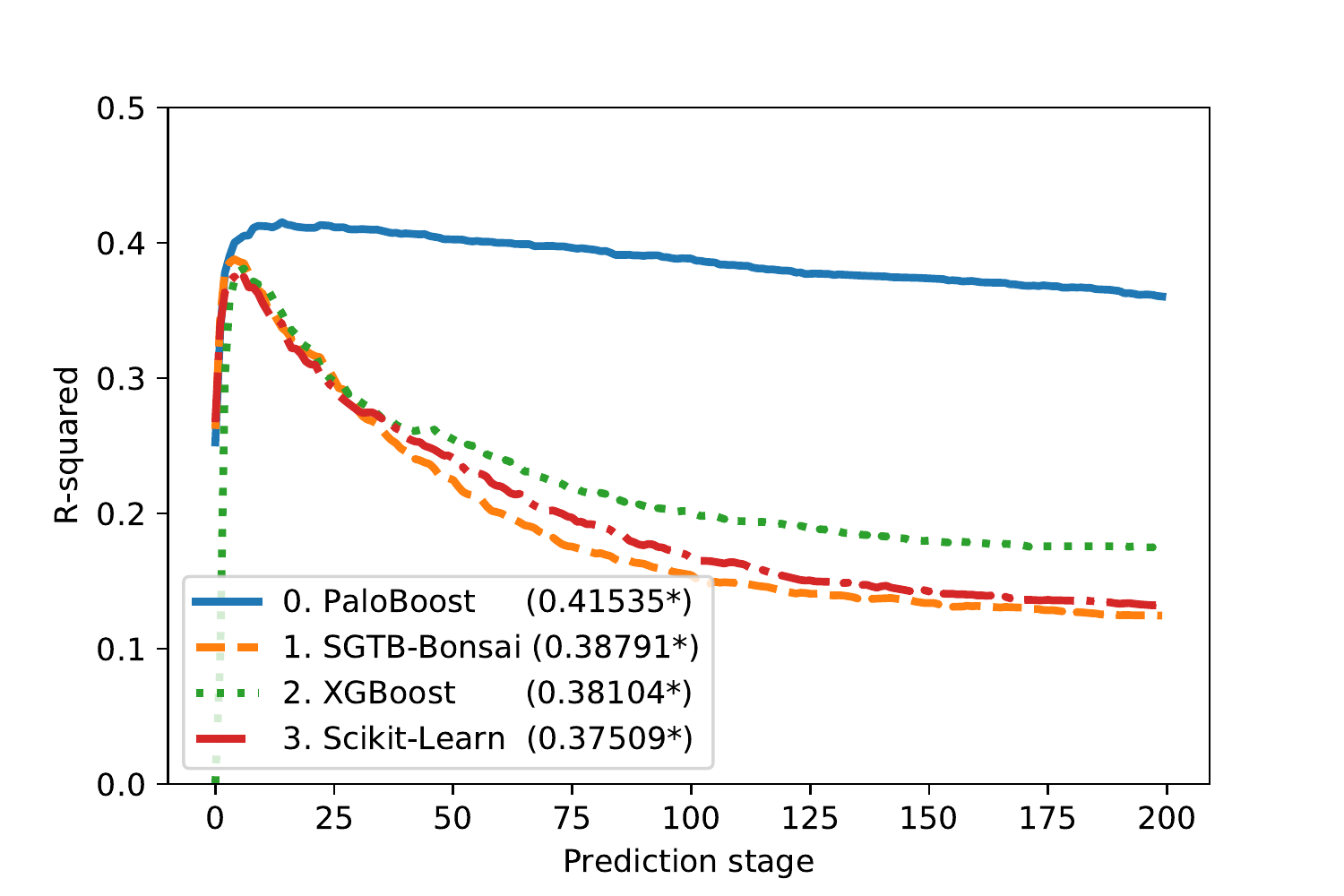}}
        \subfloat[$R^2$, $\nu_{max}=0.1$]{\includegraphics[width=0.33\textwidth]{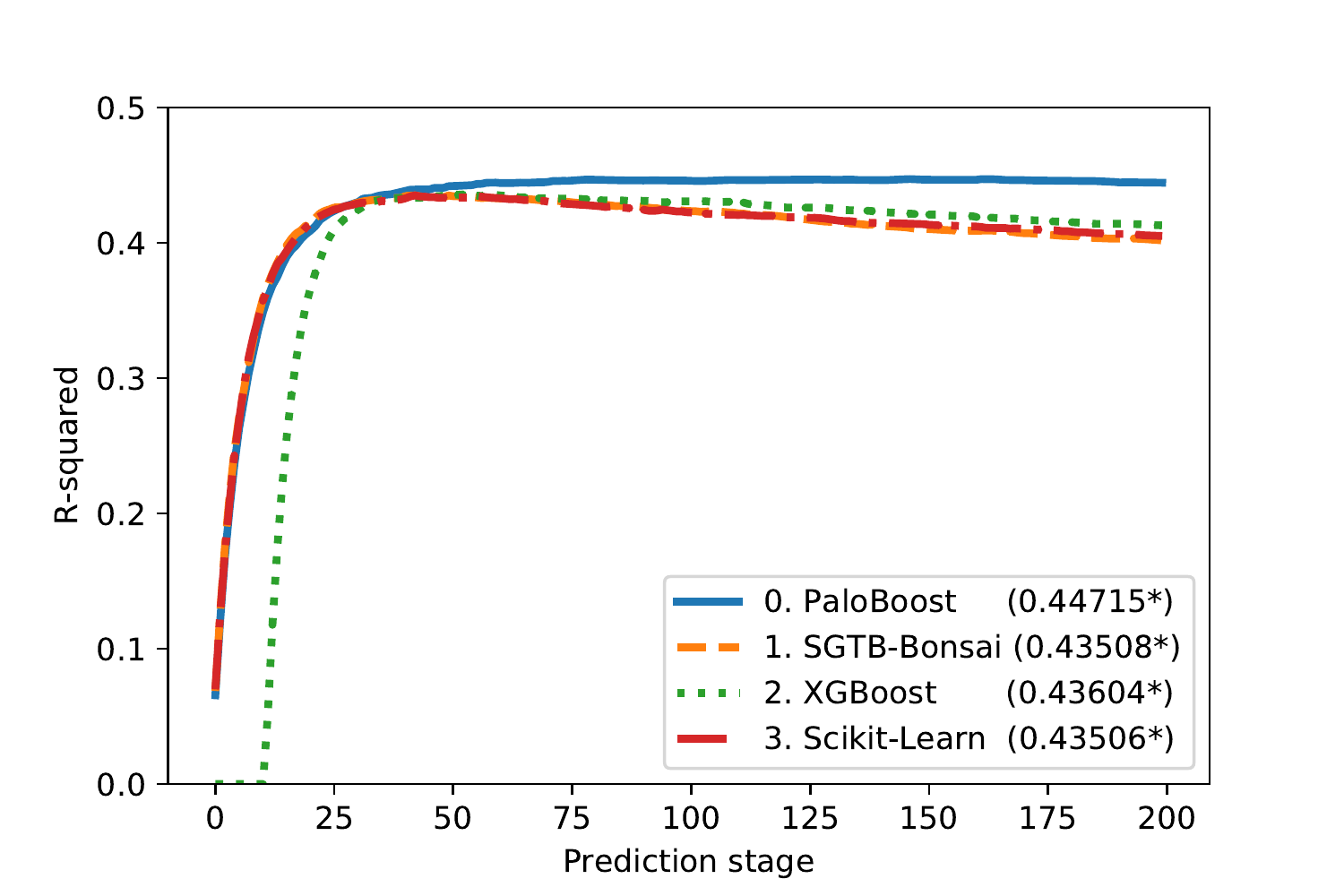}}\\
        \caption{Predictive performance on the Simulated Dataset.}
    \label{fig:friedman-perf}
\end{figure}

\emph{Predictive Performance}. Figure \ref{fig:friedman-perf} shows the coefficient of determination ($R^2$) for the three different learning rates.
PaloBoost exhibits the best and most stable predictive performance for all the learning rates.
While the other SGTB implementations show a significant degradation in predictive performance as they iterate more, PaloBoost displays a graceful drop of predictive performance even with high learning rates.
Even for a small learning rate of 0.1, the overfitting behavior of SGTB becomes apparent after 25 iterations.
Moreover, the predictive performance of PaloBoost varies considerably less across the three different learning rates, illustrating better robustness to the parameter setting ($\nu_{max}$).

\begin{figure}[tb]
\centering
        \subfloat[Avg. LR, $\nu_{max}=1.0$]{\includegraphics[width=0.33\textwidth]{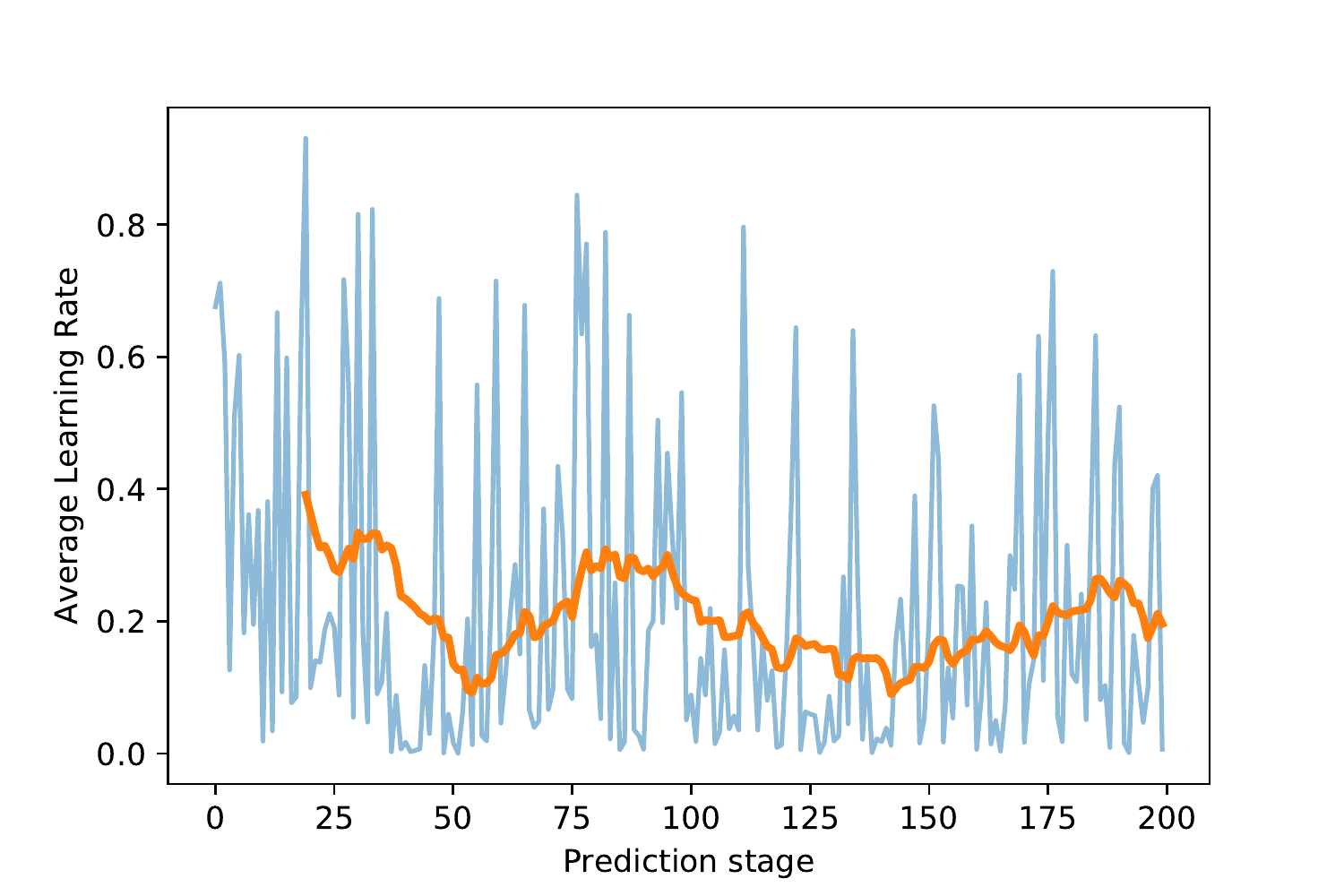}}
        \subfloat[Avg. LR, $\nu_{max}=0.5$]{\includegraphics[width=0.33\textwidth]{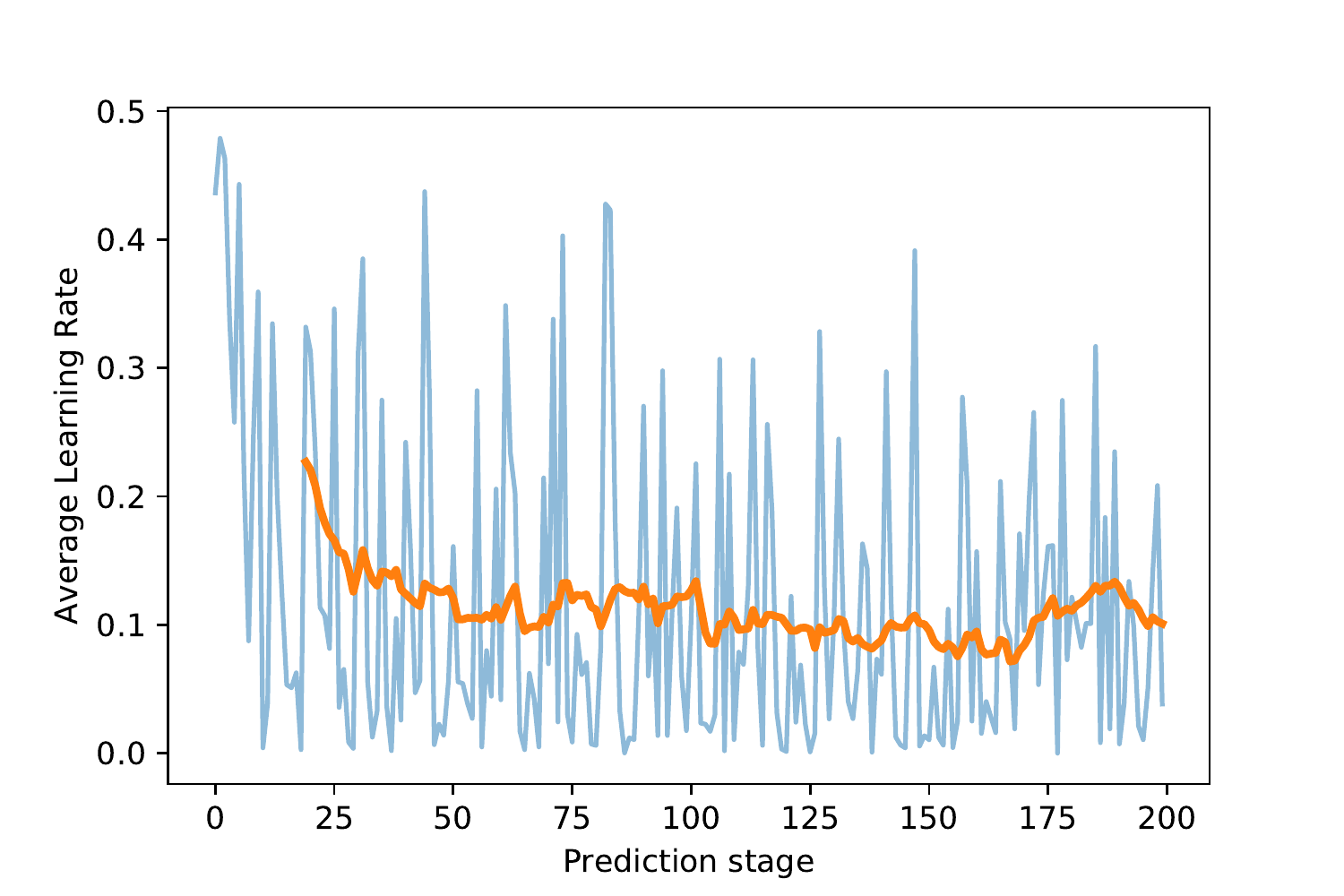}}
        \subfloat[Avg. LR, $\nu_{max}=0.1$]{\includegraphics[width=0.33\textwidth]{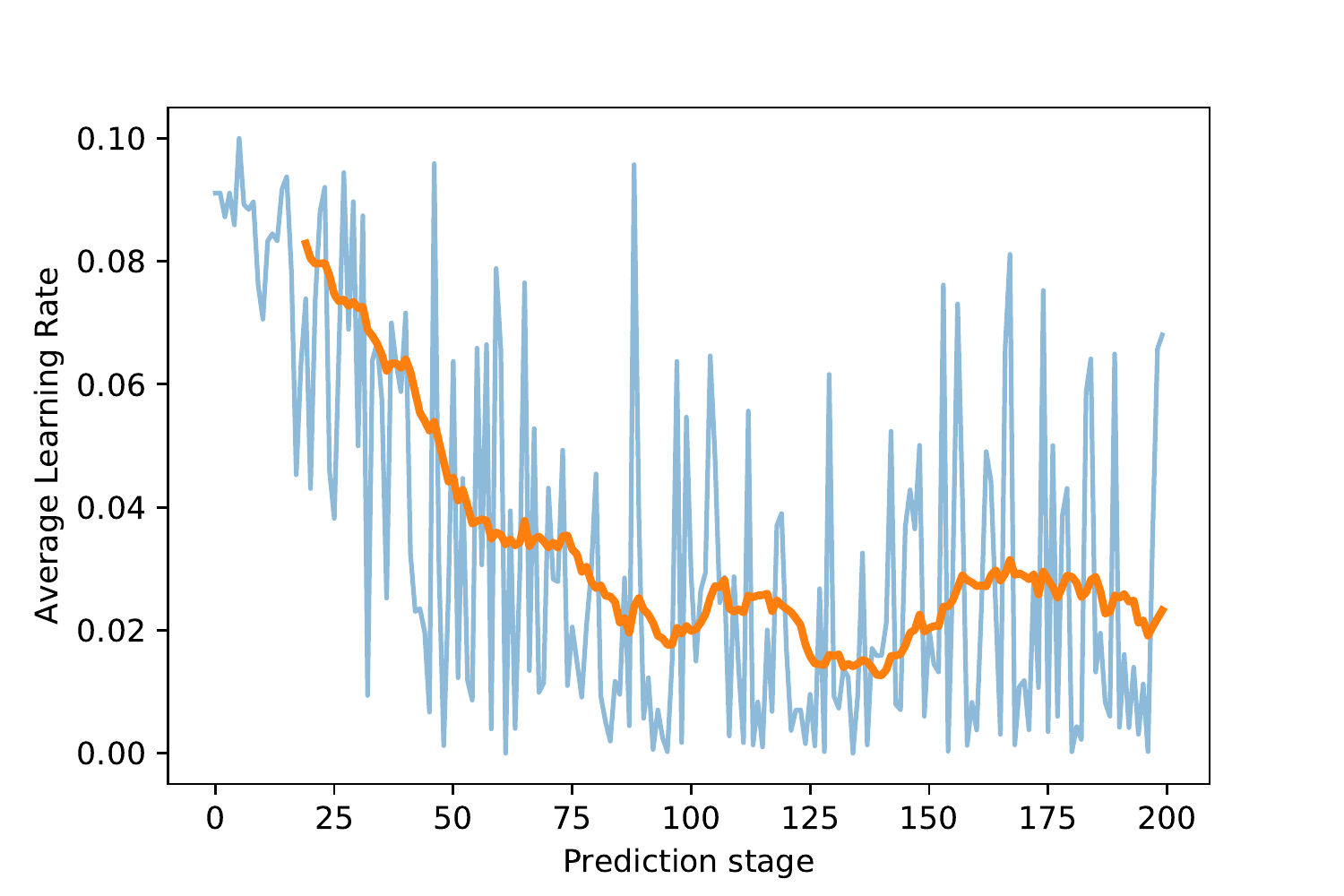}}\\
        \subfloat[Prune Rate, $\nu_{max}=1.0$]{\includegraphics[width=0.33\textwidth]{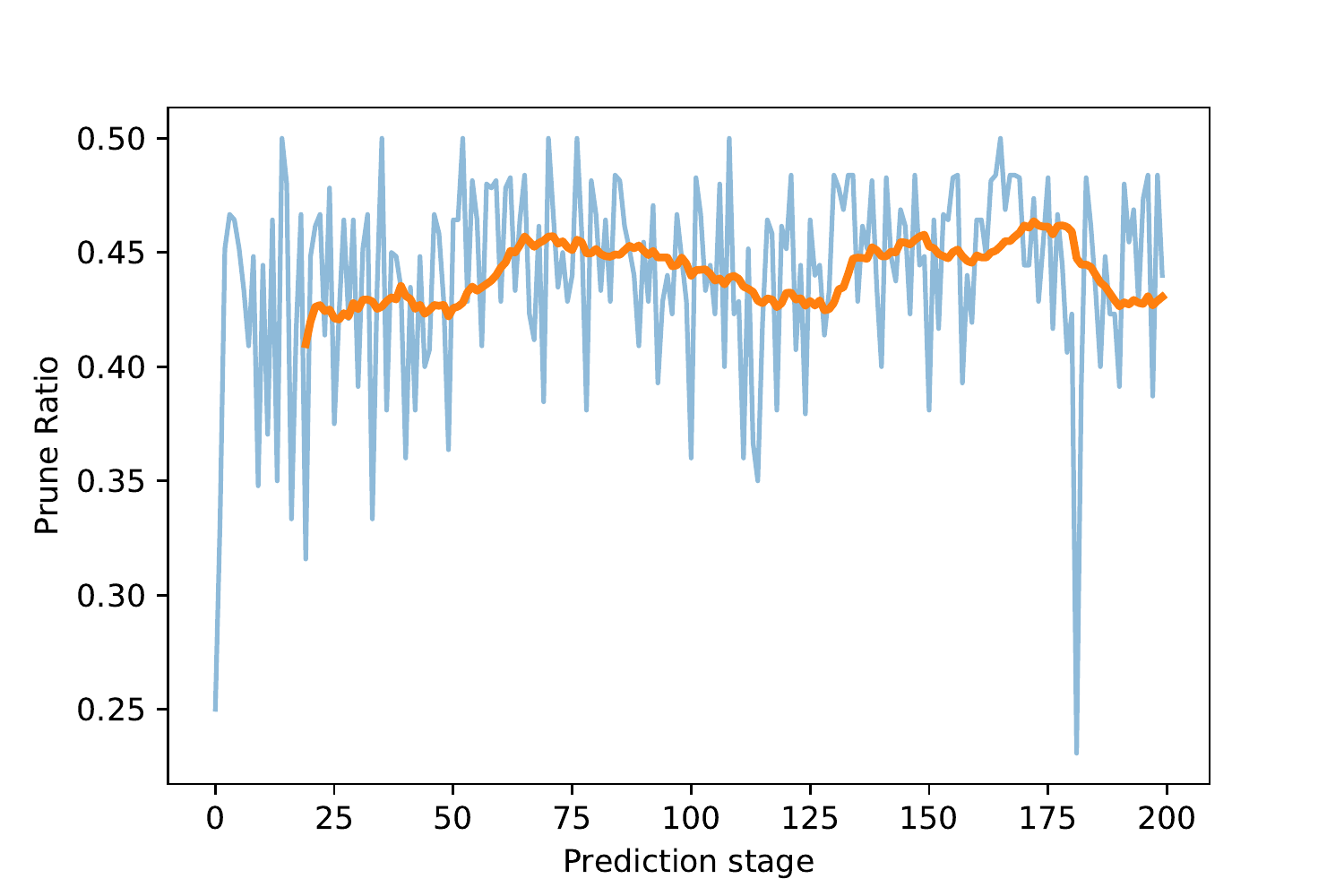}}
        \subfloat[Prune Rate, $\nu_{max}=0.5$]{\includegraphics[width=0.33\textwidth]{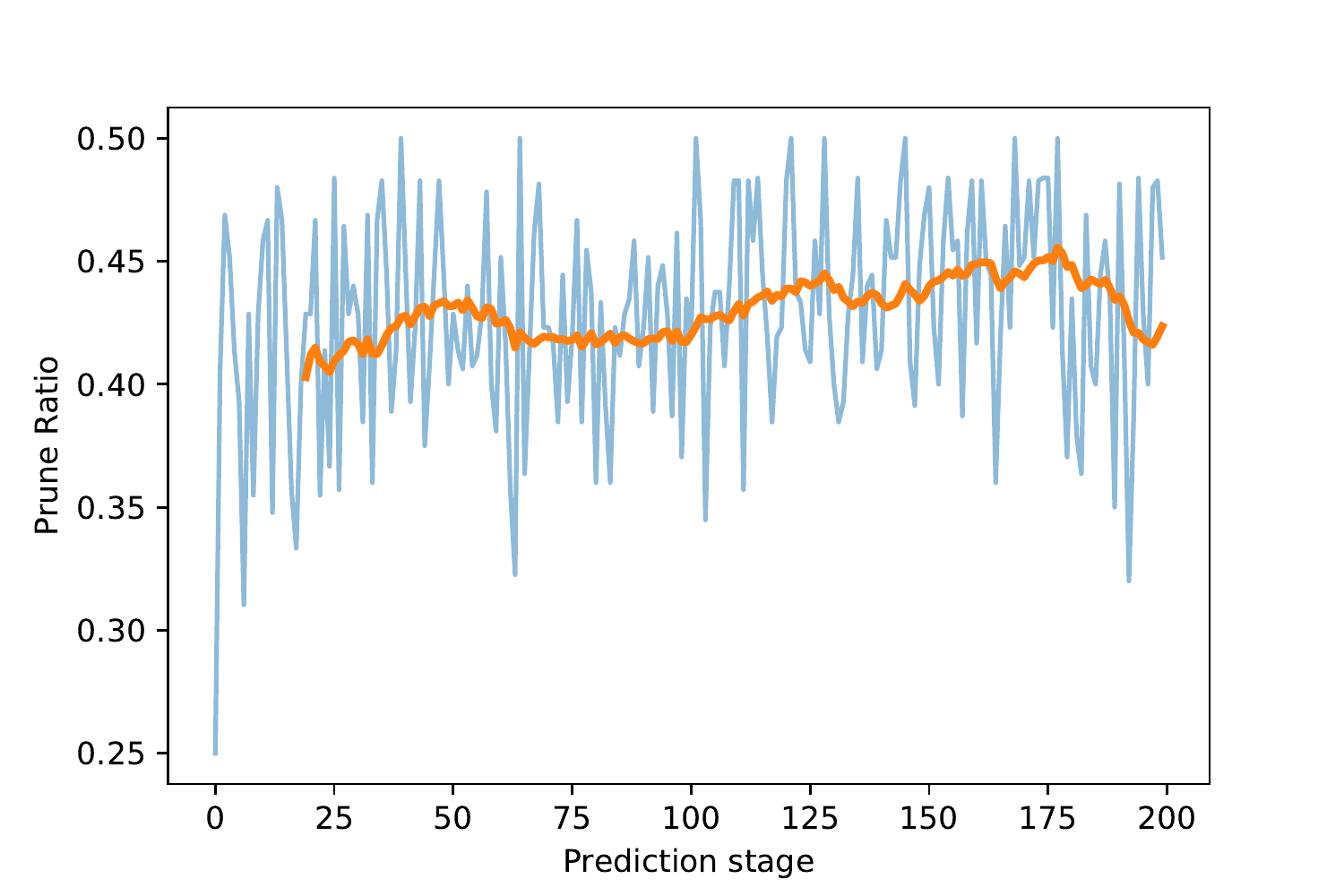}}
        \subfloat[Prune Rate, $\nu_{max}=0.1$]{\includegraphics[width=0.33\textwidth]{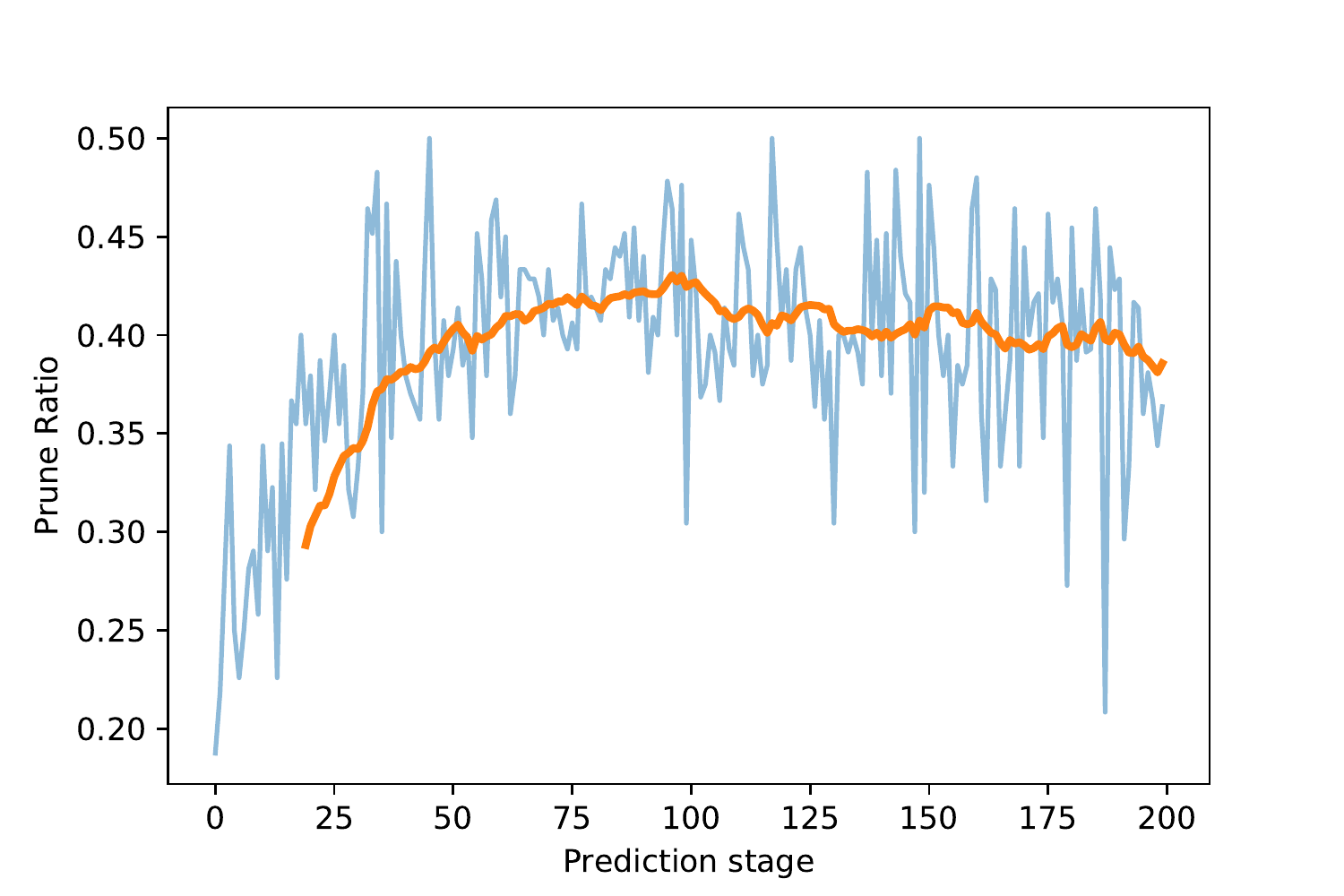}}
                \caption{The average learning rate and prune rate for PaloBoost on the Simulated Dataset.}
    \label{fig:friedman-palo}
\end{figure}

\emph{Average Learning Rate and Prune Rate}.
The gradual decline of predictive performance in PaloBoost can be better understood by analyzing the prune rate and average learning rate at each stage.
Figure \ref{fig:friedman-palo} displays the average learning rate (top row) and the prune rate (bottom row) over the same iterations.
We also overlay the 20-step moving average (colored in orange), to visualize the overall trend.
As can be seen, the learning rate for each stage is quite different, and often below the specified maximum rate.
We can also observe that as the number of stages increases, the learning rates are adaptively adjusting to smaller values (more noticeable in $\nu_{max} = 0.1$).
In addition, the prune rate increases to yield lower variance trees.
Thus, the two regularization mechanisms serve as a guard against overfitting.

\begin{figure}
    \centering
        \includegraphics[width=0.6\columnwidth]{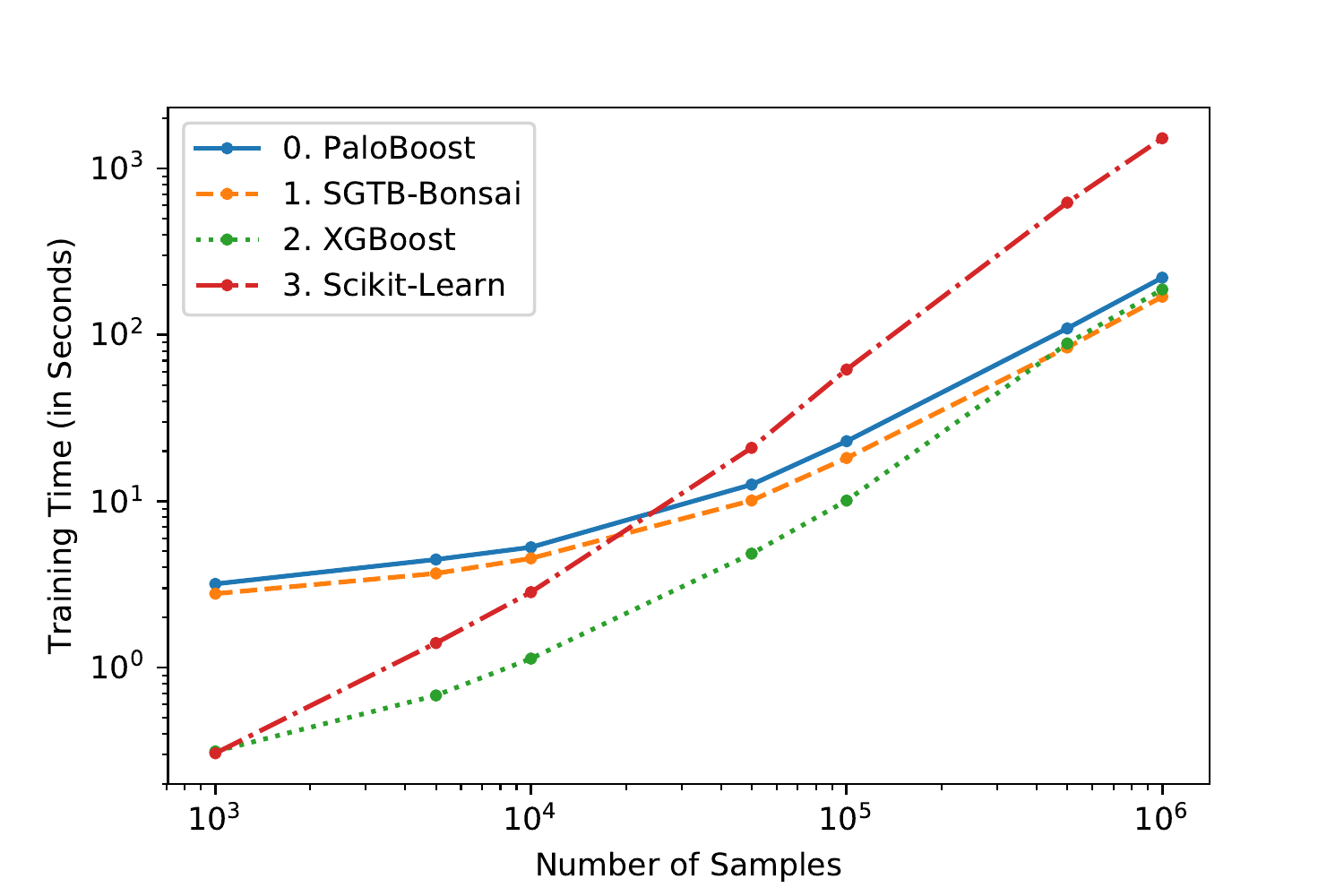}
        \caption{Runtime experiment on a macOS High Sierra 10.13.6 machine with 2.7 GHz Intel Core i5 CPU, Python 2.7.14, Scikit-Learn 0.19.1, and XGBoost 0.6.}
    \label{fig:speed}
\end{figure}

\emph{Computational Speed}.
Gradient-aware pruning and adaptive learning add computational complexity to SGTB. 
To determine the impact of our modifications, we measured the training speeds of the four SGTB models (PaloBoost, Bonsai-SGTB, Scikit-learn, and XGBoost). 
Figure~\ref{fig:speed} shows the measured training time over simulated training sample sizes.
As can be seen, the performance of PaloBoost is comparable with the other implementations even for large sample sizes (e.g. one million samples).
While PaloBoost is marginally slower than SGTB-Bonsai, the two regularization techniques do not slow down the overall training process much.

\begin{figure}
    \centering
        \includegraphics[width=1.0\textwidth]{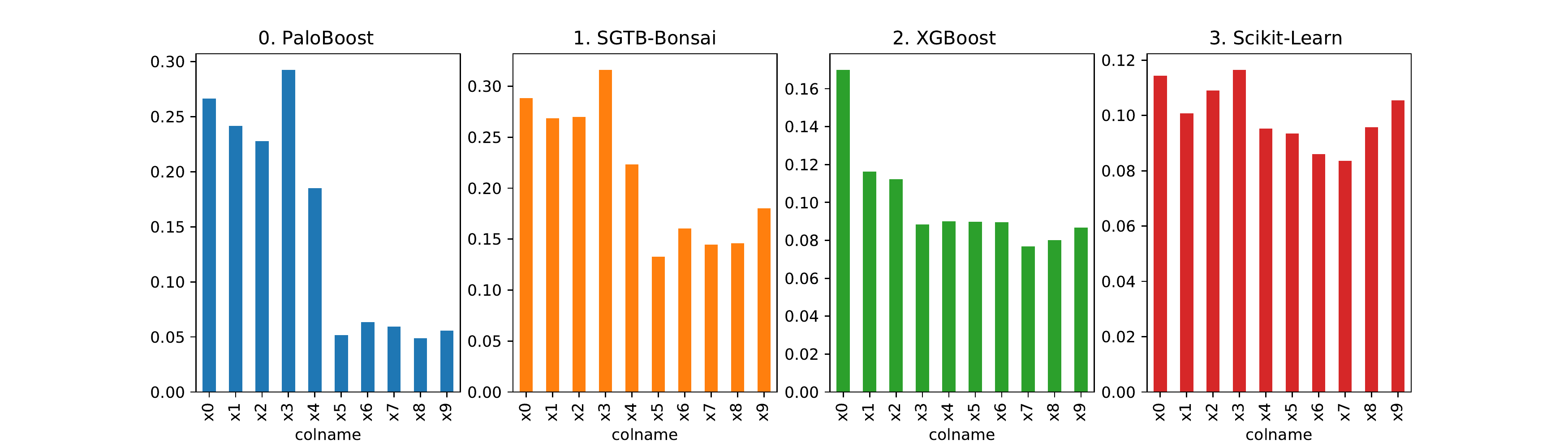}
        \caption{The feature importances after 200 iterations. Only the first five features, $x_0$ to $x_4$, are used to generate the target.}
    \label{fig:feature-importance}
\end{figure}

\begin{figure}
    \centering
            \includegraphics[width=0.6\textwidth]{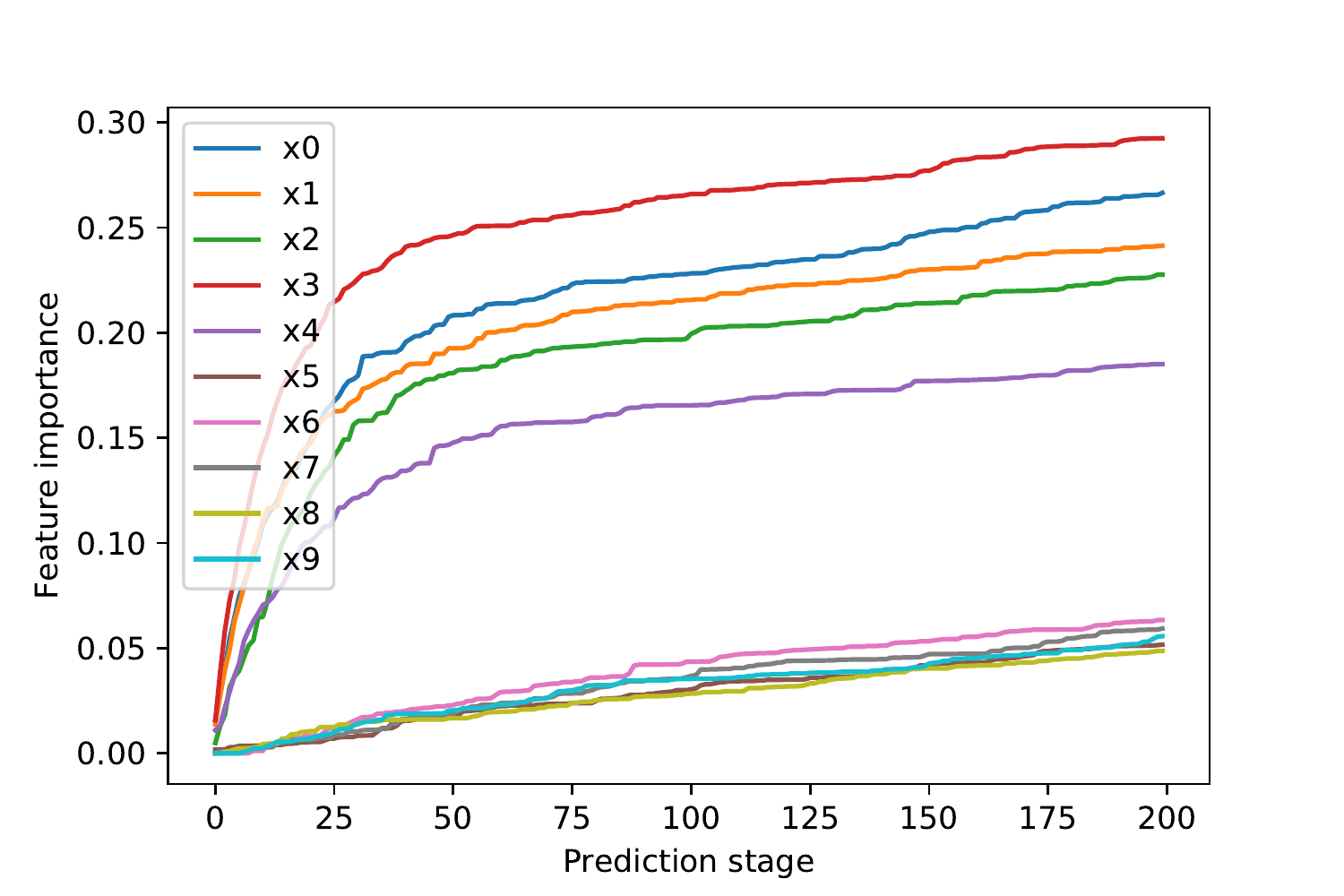}
        \caption{The evolution of feature importances at each stage of PaloBoost. This plot resembles the coefficient path plot in Elastic Net \cite{Zou:2005}.}
    \label{fig:feature-importance-stage}
\end{figure}

\emph{Feature Importance}.
Since the dataset is simulated, the true feature importance is known, with only 5 features ($x_0 - x_4$) used to generate the target.
We compared the feature importances from the four different trained models with 200 iterations ($M=200$) and a learning rate of 0.1 ($\nu_{max}=0.1$).
PaloBoost and SGTB-Bonsai use the proposed feature importance formula (see Section \ref{sec:paloboost-feaures}), while XGBoost and Scikit-Learn use the standard feature importance formula (summation of the squared error improvements per feature).
Figure~\ref{fig:feature-importance} shows the feature importances from all four SGTB models\footnote{The results are similar for $M=50$ but less dramatic.}.
PaloBoost clearly identifies the noisy features ($x_5-x_9$), while XGBoost and Scikit-Learn estimate similar importances between the relevant and noisy features.
Although preliminary, this indicates PaloBoost can be also very useful for the feature selection processes.

The superiority of PaloBoost's feature importance is likely due to two different aspects.
First, gradient-aware pruning and adaptive learning rates are removing noisy features, as regions that do not generalize well are either pruned or have a small adaptive learning rate.
Secondly, the modified importance formula encapsulates the learning rate, coverage, and impact of the feature, using the leaf-centric multiplicative factor $\nu_{jm} |R_{jm}|  |\gamma_{jm}|$.
This can be confirmed in Figure \ref{fig:feature-importance} by comparing the importance estimates for features $x_5-x_9$ between SGTB-Bonsai and XGBoost (or Scikit-Learn).
SGTB-Bonsai uses the proposed feature importance formula with the same learning rate ($nu_{jm} = 0.1$), yet still identifies lower importance values for $x_5-x_9$.
While accounting for the region size ($|R_{jm}|$) and node estimate ($\gamma_{jm}$) is clearly important, the adaptive learning rate provides further separation of the noisy features.
Thus, we conclude that both aspects play important roles for the superior performance.

We further explored PaloBoost's feature importance estimates as a function of the iterations for the same learning rate ($\nu_{max}=0.1$).
Figure~\ref{fig:feature-importance-stage} visualizes the evolution of the feature importance estimates and offers a perspective analogous to the coefficient path in Elastic Net \cite{Zou:2005}.
Using this plot, we can inspect the inner mechanism of PaloBoost more thoroughly.
As can be seen, the feature importances for the first five features rapidly increases through the first 30 iterations, after which they slowly increase.
Moreover, after 20 iterations, the order of the features no longer change and reflect the true feature rank.
We also note that this visualization can also be used to determine the stopping criteria for the number of iterations.

\begin{figure*}
    \centering
        \subfloat[Mercedes, $\nu_{max}=1.0$]{\includegraphics[width=0.33\textwidth]{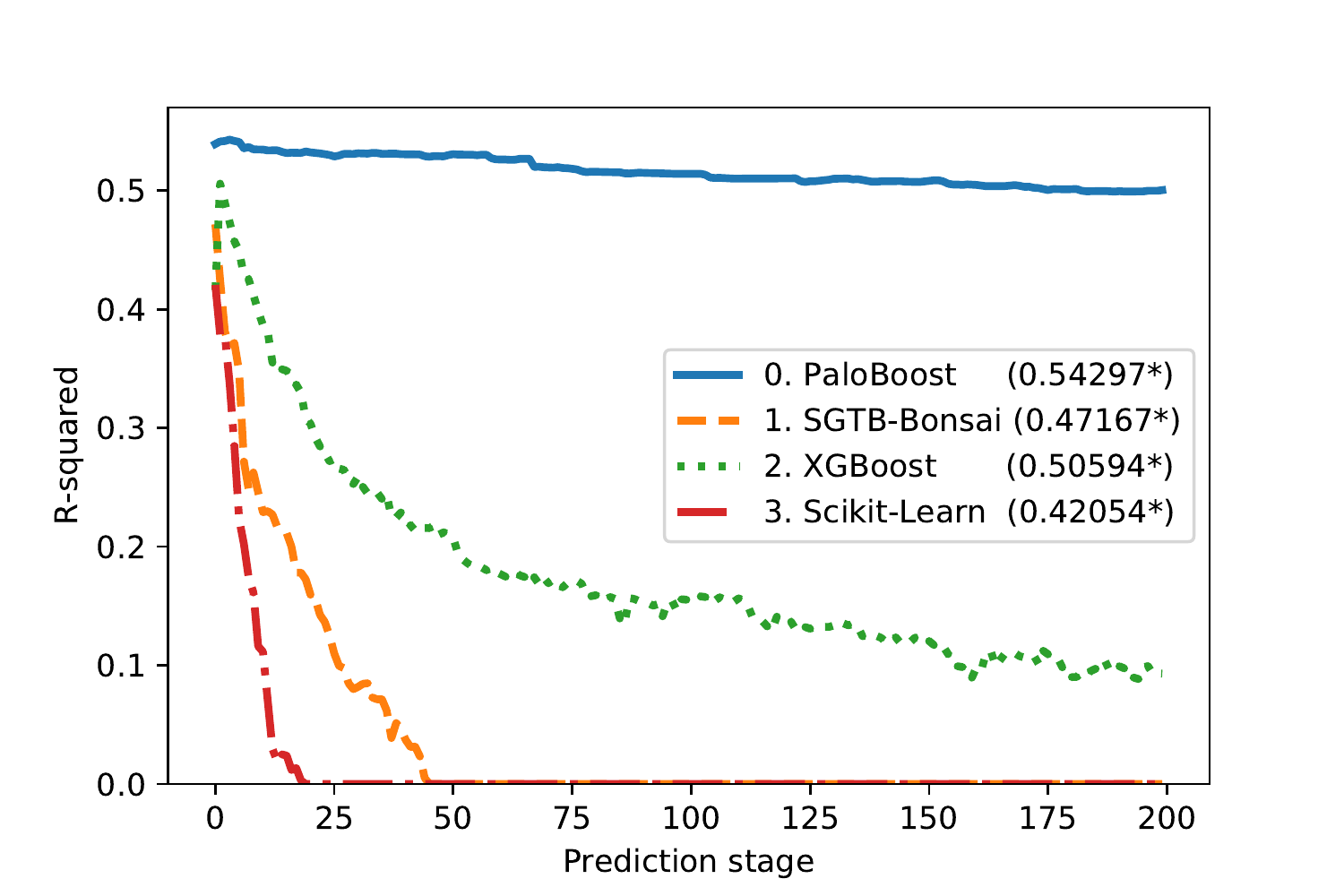}}
        \subfloat[Mercedes, $\nu_{max}=0.5$]{\includegraphics[width=0.33\textwidth]{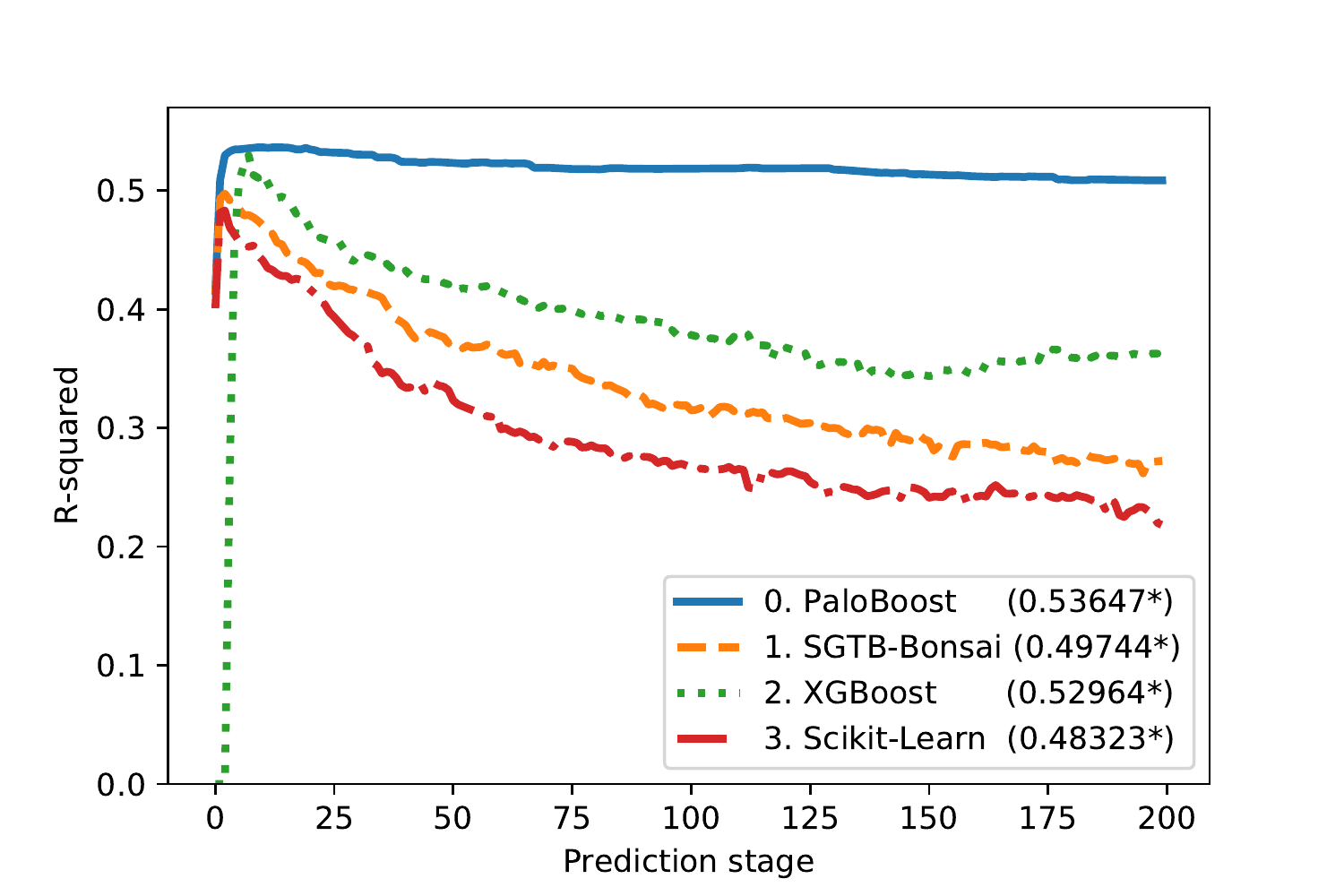}}
        \subfloat[Mercedes, $\nu_{max}=0.1$]{\includegraphics[width=0.33\textwidth]{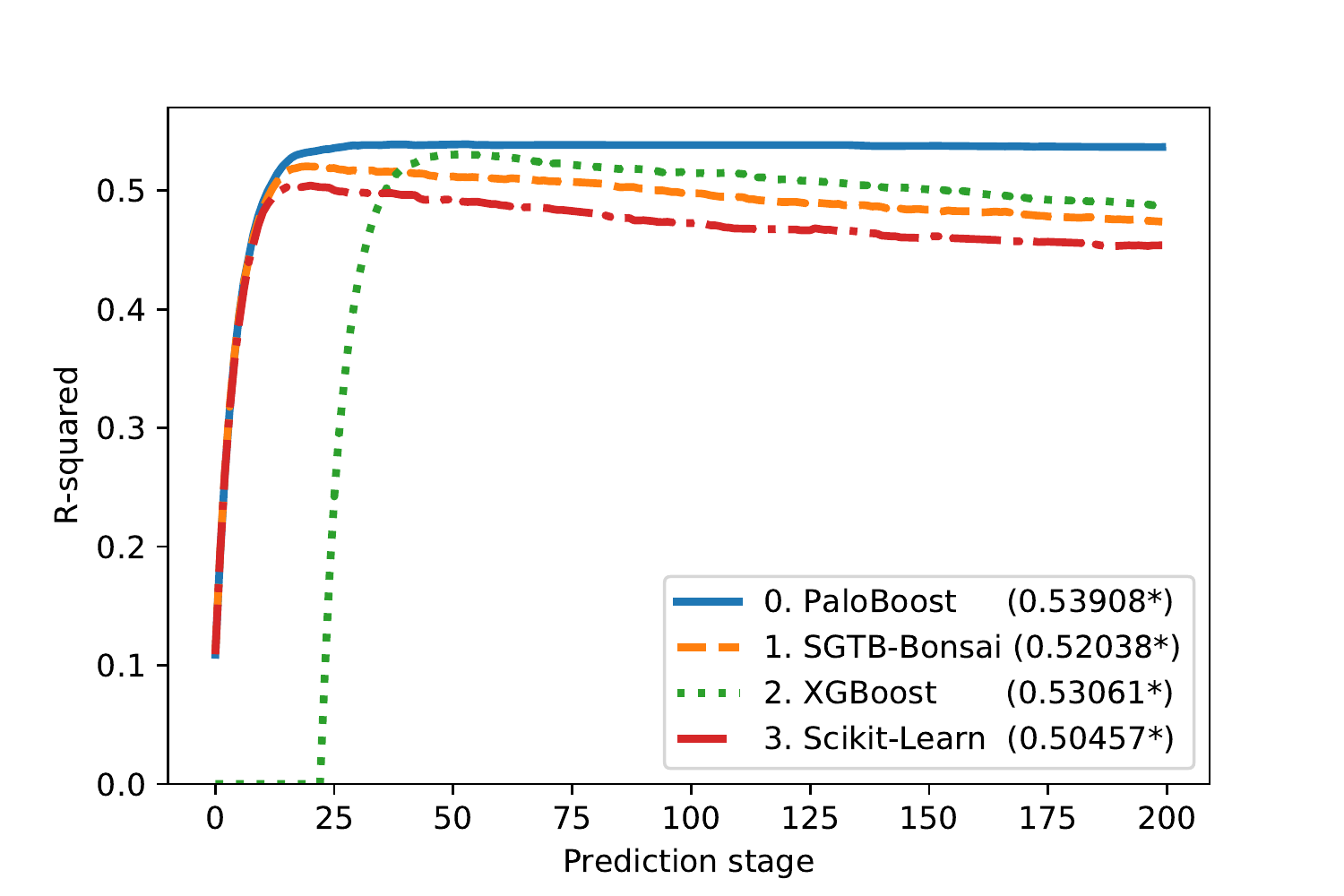}} \\
        \subfloat[Community, $\nu_{max}=1.0$]{\includegraphics[width=0.33\textwidth]{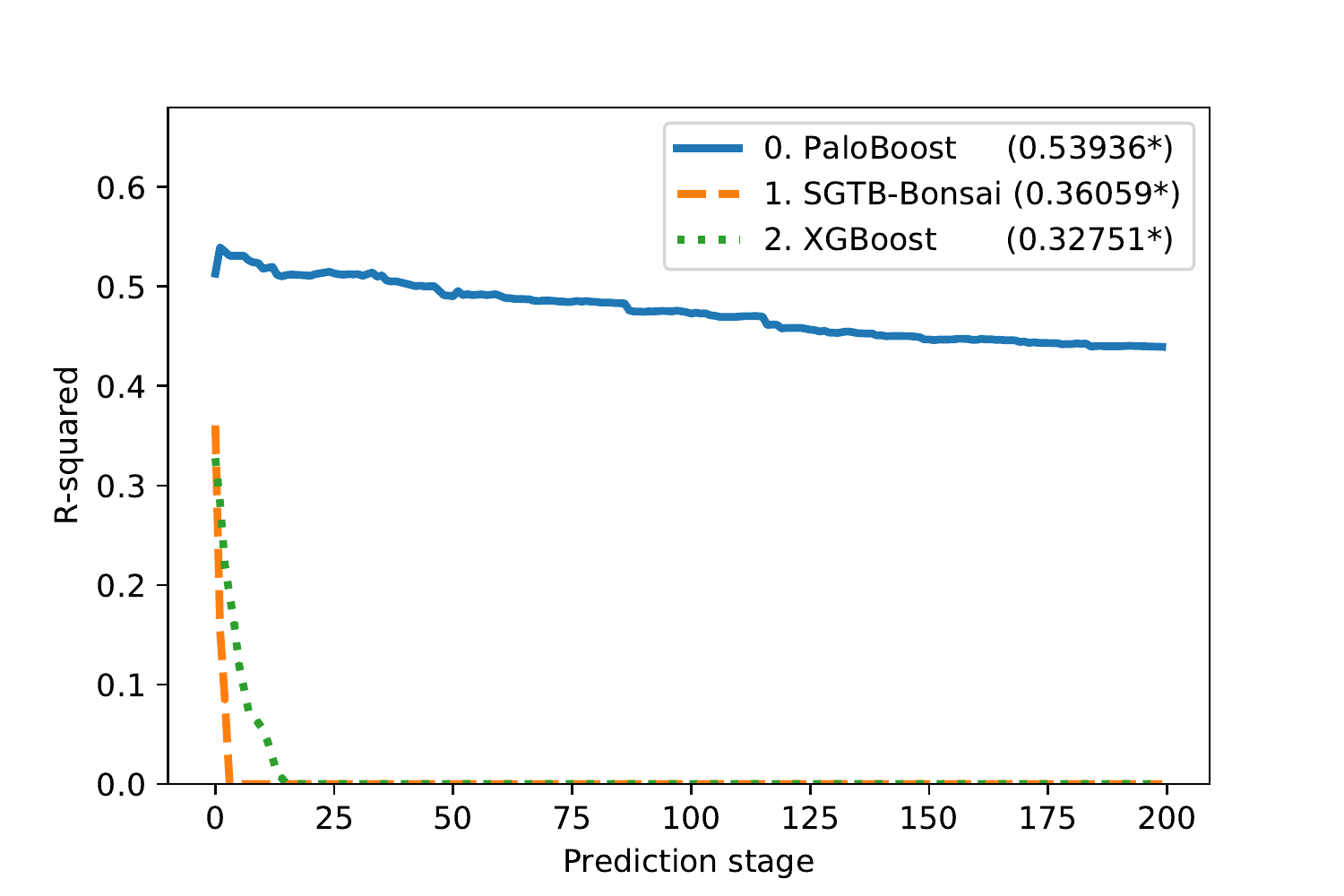}}
        \subfloat[Community, $\nu_{max}=0.5$]{\includegraphics[width=0.33\textwidth]{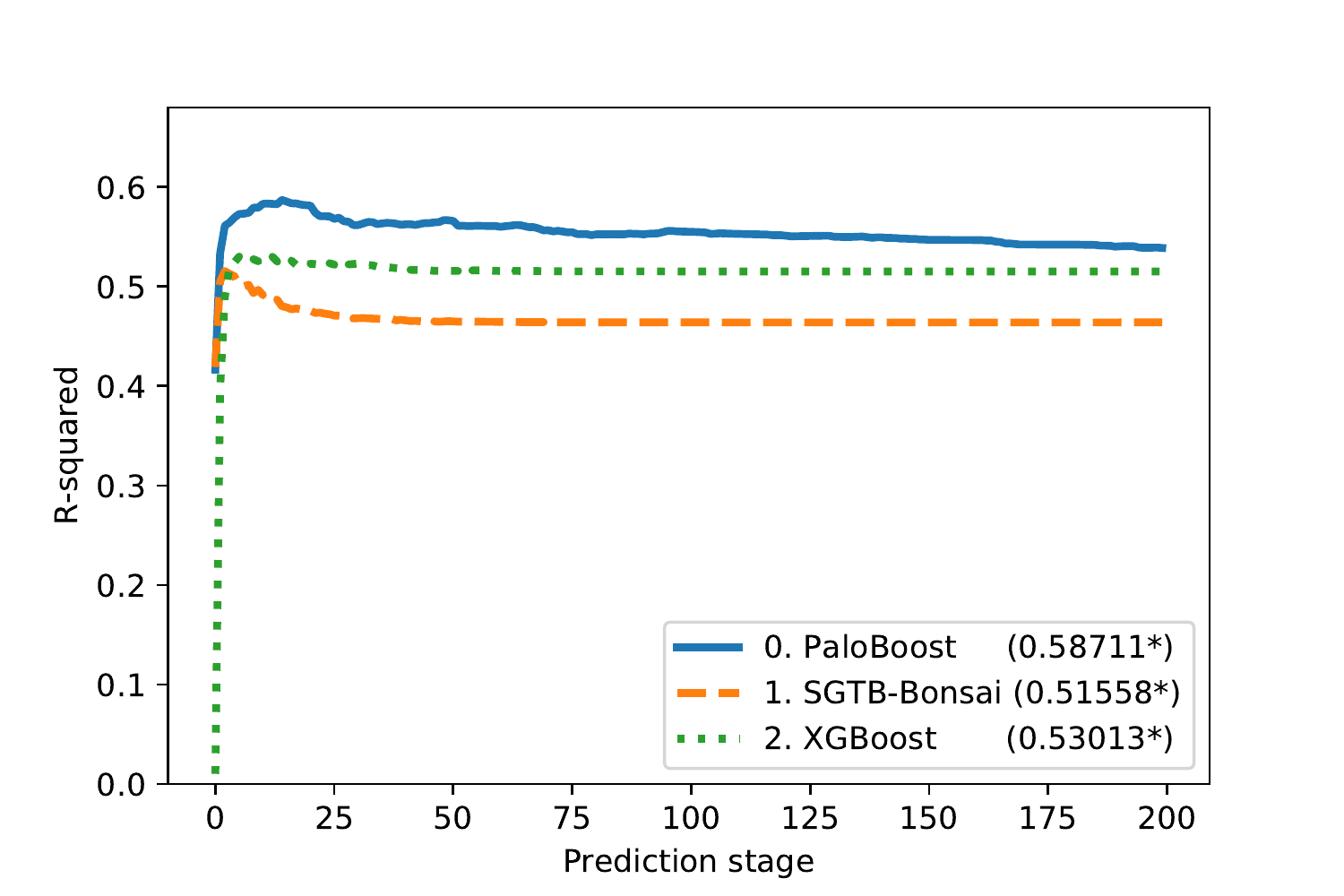}}
        \subfloat[Community, $\nu_{max}=0.1$]{\includegraphics[width=0.33\textwidth]{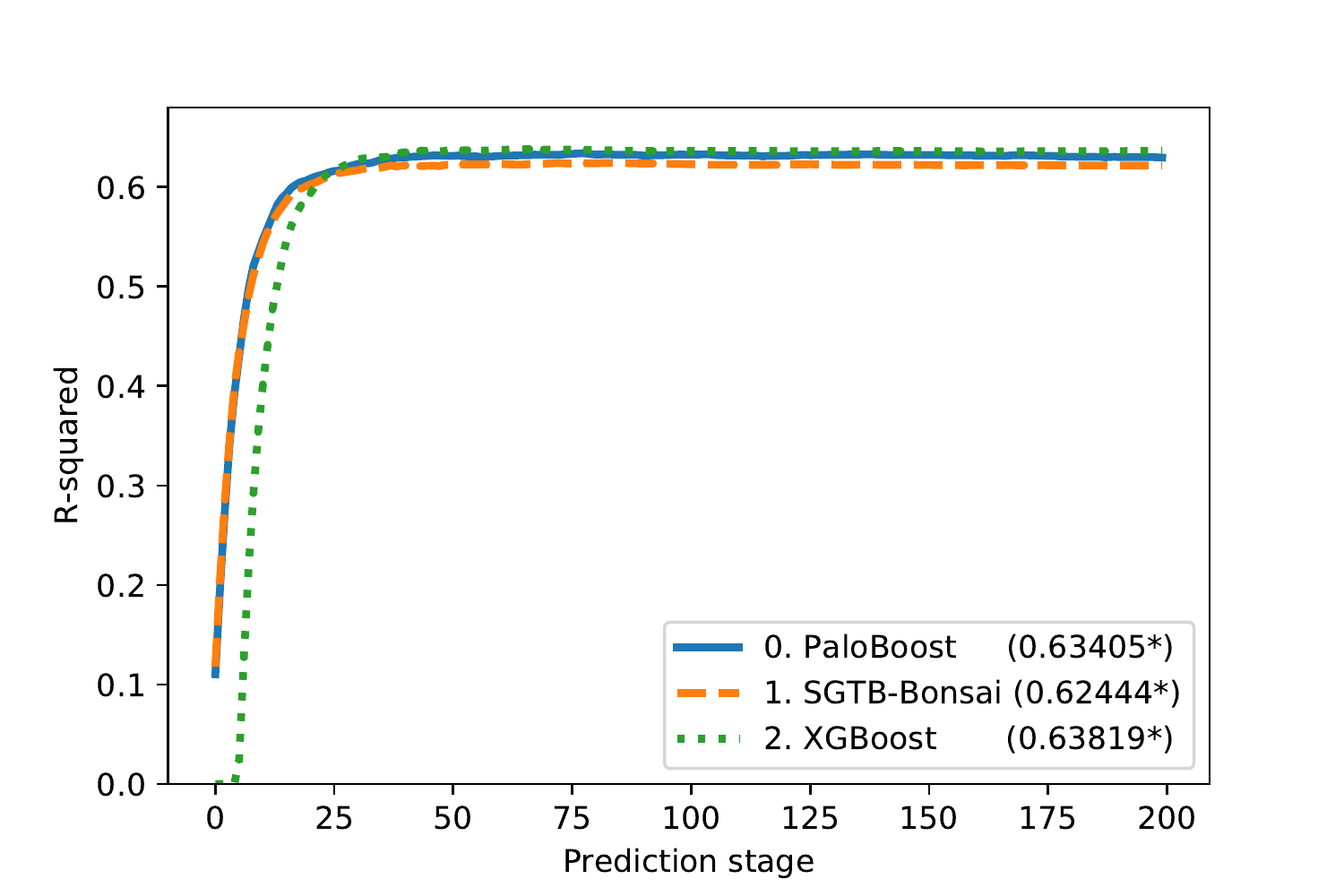}}\\
        \caption{The predictive performance, measured by $R^2$, on the regression tasks.
                    The results from Mercedes-Benz dataset are on the top row, and the Community-Crime on the bottom row. 
              The y-axis is aligned for each dataset (row) for ease of comparison across learning rates.
    }
    \label{fig:regression}
\end{figure*}

\subsection{Regression Tasks}

The four SGTB models are evaluated on two publicly available regression datasets, the Mercedes-Benz Greener Manufacturing Dataset and the Communities Crime Dataset.
While only the coefficient of determination ($R^2$) results are presented, the results from our extensive experimental studies on adaptive learning rate and prune rate are available on the PaloBoost package webpage.

\subsubsection{Mercedes-Benz Greener Manufacturing Dataset}
Daimler, the parent company of Mercedes-Benz, performs exhaustive tests on its pre-release cars for the safety and reliability.
The process can easily consume a substantial amount of gas and time, and generate a large volume of carbon dioxide.
In the effort to reduce the testing time and conserve the environment, 
the company released its dataset that contains various measurements (features) and time to finish the test (target) through the Kaggle platform\footnote{https://www.kaggle.com/c/mercedes-benz-greener-manufacturing}.
As summarized in Table \ref{tab:data}, the dataset has only 4,209 samples with 460 features (post-processed). 
Thus, many complex models are likely to overfit on the small training sample and need to properly address the curse of dimensionality \cite{Trunk:1979, Hughes:1968}.

The top row in Figure~\ref{fig:regression} shows the predictive performance ($R^2$) on the Mercedes-Benz dataset.
Similar to the simulation results (Figure \ref{fig:friedman-perf}), PaloBoost achieves the best and most stable performance.
More impressively, PaloBoost suffers minimal degradation at the high learning rate ($\nu_{max} = 1.0$), while a noticeable decay occurs immediately for the other three models.
Furthermore, the performance differences across the three learning rates are marginal for PaloBoost, which demonstrates the robustness of the maximum learning rate parameter setting.
Unlike the other SGTB implementations that use a fixed learning rate, we observed that the average learning rate in PaloBoost dropped after a few iterations.
Moreover, approximately 40\% of the regions were pruned during the gradient-aware prune mechanism.
Together, these modifications contribute to the robust performance of PaloBoost.

\subsubsection{Communities and Crime Dataset}
Communities that have similar socioeconomic and law enforcement profiles may also experience similar types of crimes. 
Such information can be shared across police departments to enable cooperation.
In the effort to make such a data-driven system, 
Redmond \cite{Redmond:2002} created a dataset by combining three data sources: US Census in 1990, US FBI Uniform Crime Report in 1995, and 1990 US Law Enforcement Management and Administrative Statistics Survey.
The dataset, available on the UCI Repository\footnote{http://archive.ics.uci.edu/ml/datasets/communities+and+crime}, contains 127 features that include the percent of unemployed, percent under poverty, median income, etc. and the violent crime rate of the community (numeric target).
This dataset contains only 1,994 samples with 127 features and has missing values (see Table \ref{tab:data}), therefore Scikit-Learn is omitted as a baseline.
The author of the dataset has stated that it has been carefully curated with only features that have a plausible connection to crime.

The bottom row in Figure~\ref{fig:regression} illustrates the results from the Communities and Crime Dataset experiments. 
As can be seen, PaloBoost exhibits the most stable behavior across many iterations.
However, unlike the previous two datasets (Mercedes-Benz and simulated), the performance of the benchmark models do not suffer a significant degradation.
Since the feature selection has already been done by the author, the features are not noisy, thereby mitigating the curse of dimensionality.
While PaloBoost does not obtain the highest $R^2$ overall compared to XGBoost, we note that the values between the different learning rates are much smaller in comparison. 
This eliminates the need to finely tune the learning rate and the number of iterations.

\subsection{Classification Tasks}

Next, we present the various SGTB models on four different classification tasks: Amazon Employee Access, Pulsar Detection, Carvana, and BNP-Paribas.
Although we only present the predictive performance (AUROC), the adaptive learning rate and prune rate results are available on the PaloBoost package webpage.

\begin{figure*}
    \centering
        \subfloat[Amazon, $\nu_{max}=1.0$]{\includegraphics[width=0.33\textwidth]{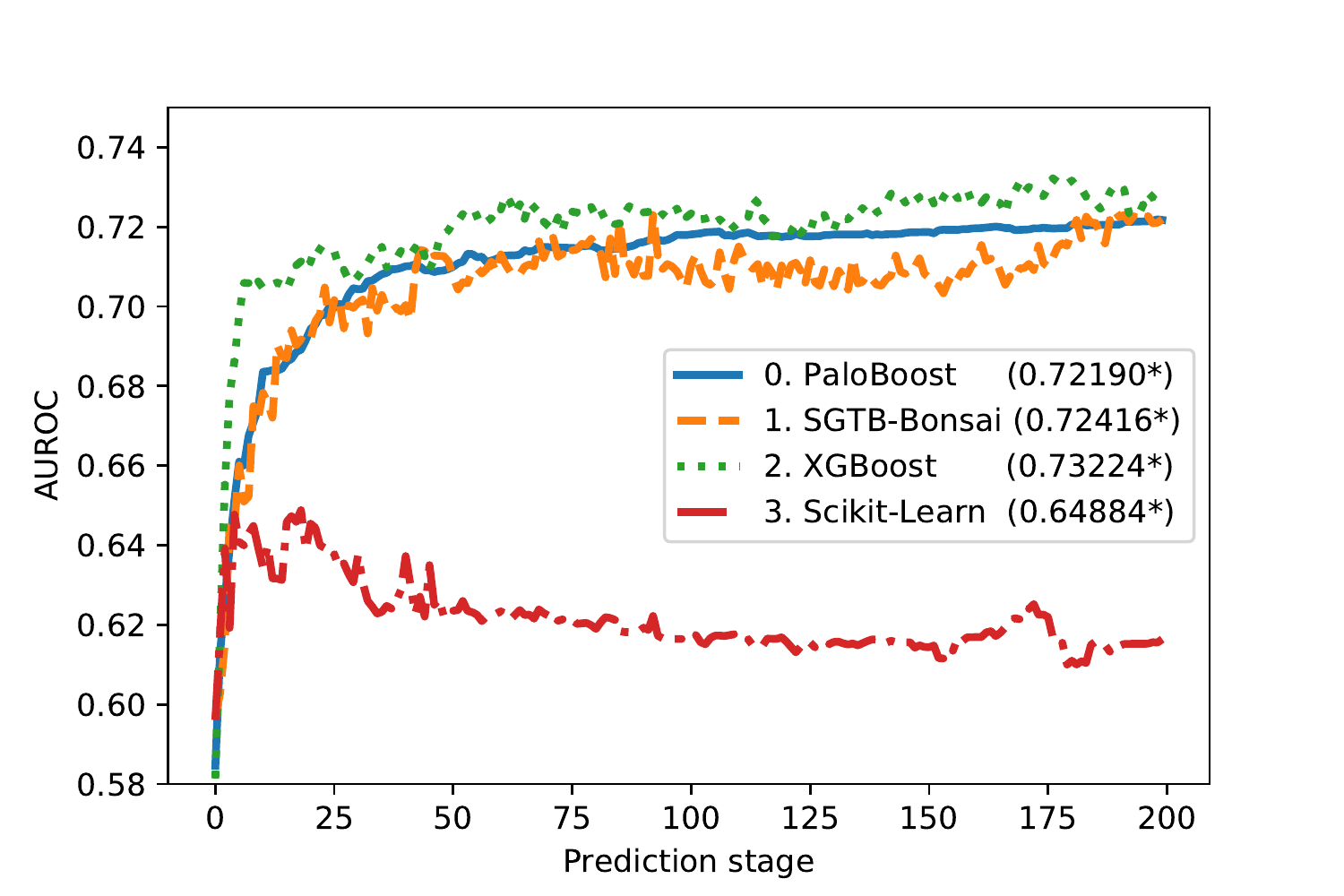}}
        \subfloat[Amazon, $\nu_{max}=0.5$]{\includegraphics[width=0.33\textwidth]{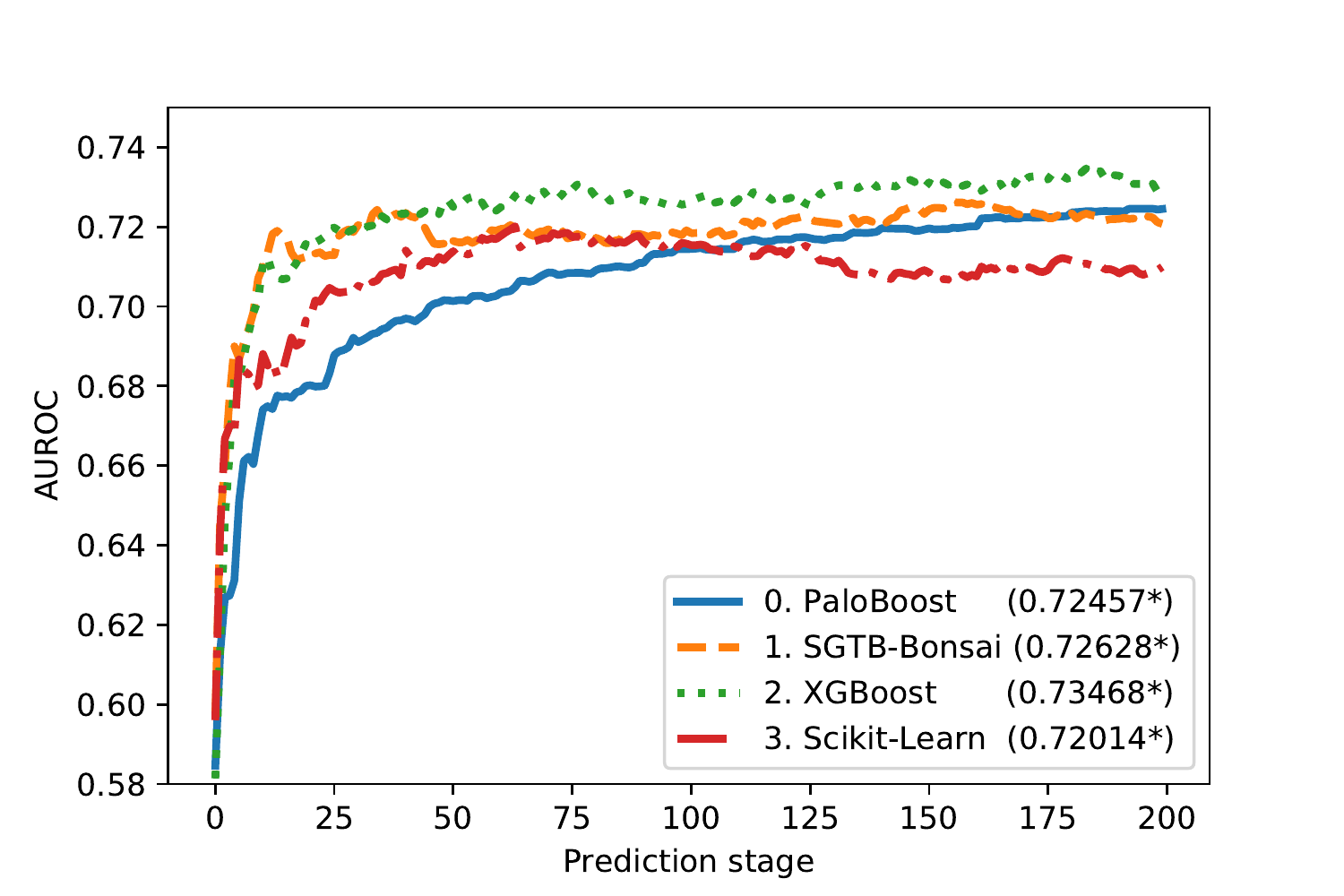}}
        \subfloat[Amazon, $\nu_{max}=0.1$]{\includegraphics[width=0.33\textwidth]{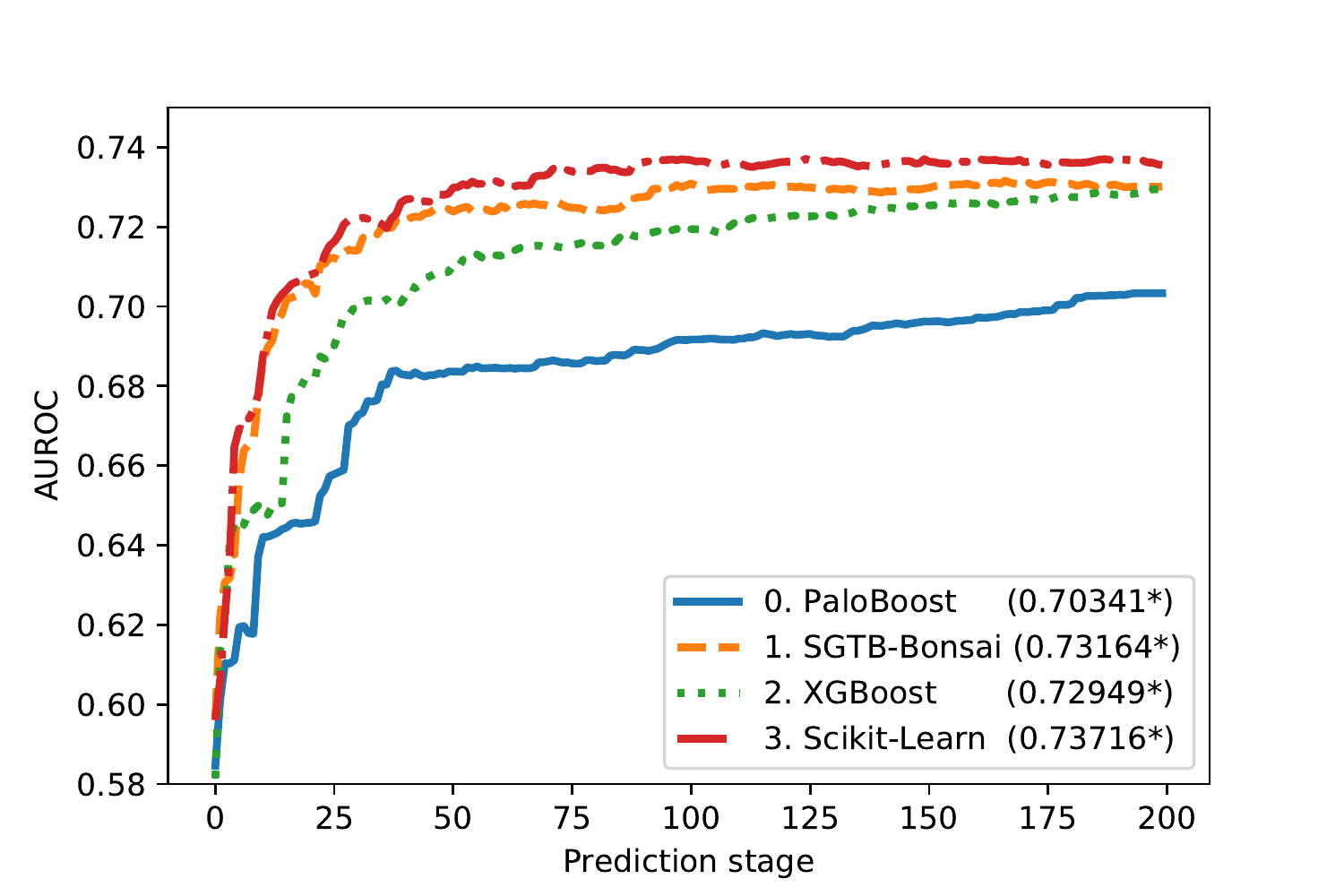}}\\
        \subfloat[Pulsar, $\nu_{max}=1.0$]{\includegraphics[width=0.33\textwidth]{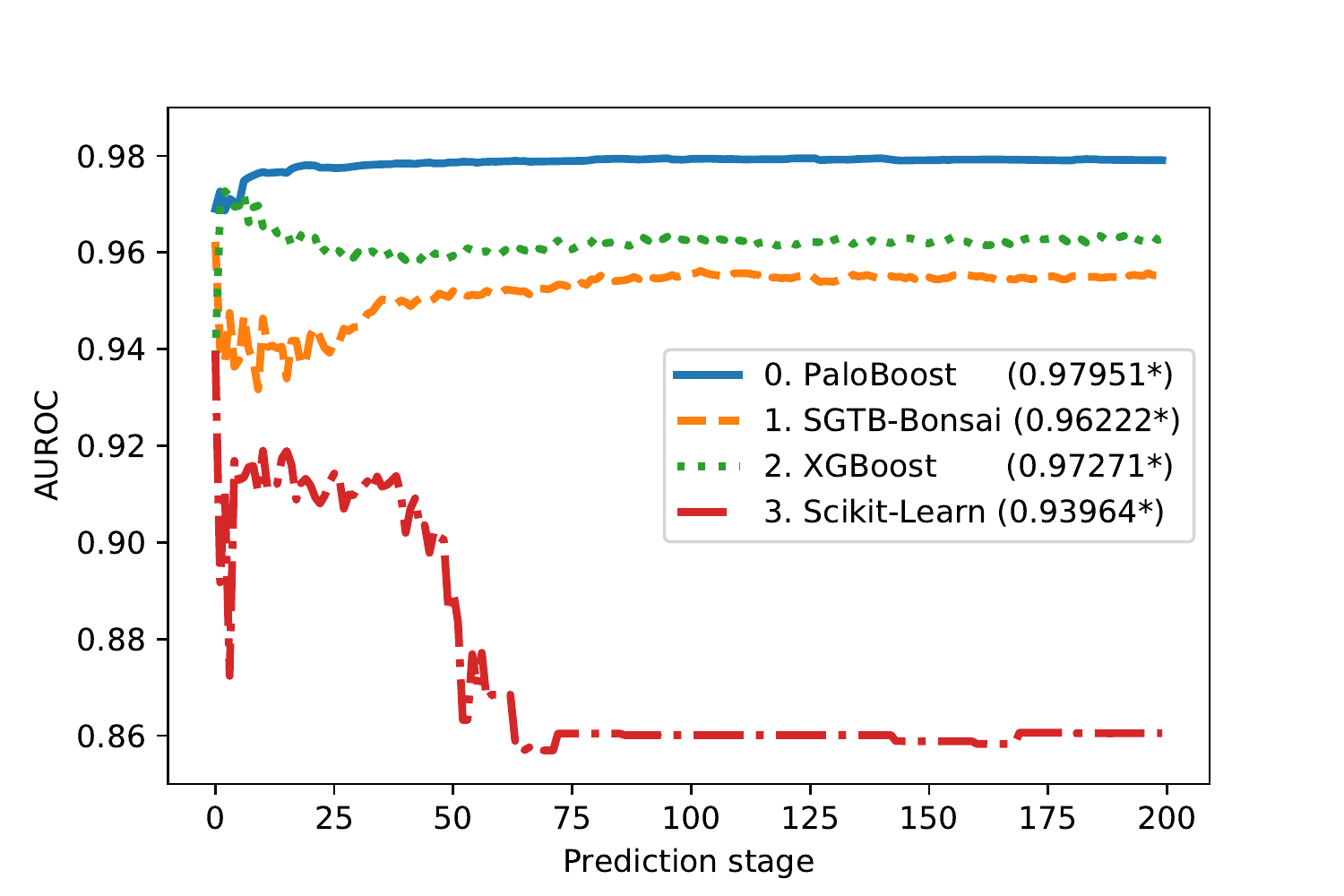}}
        \subfloat[Pulsar, $\nu_{max}=0.5$]{\includegraphics[width=0.33\textwidth]{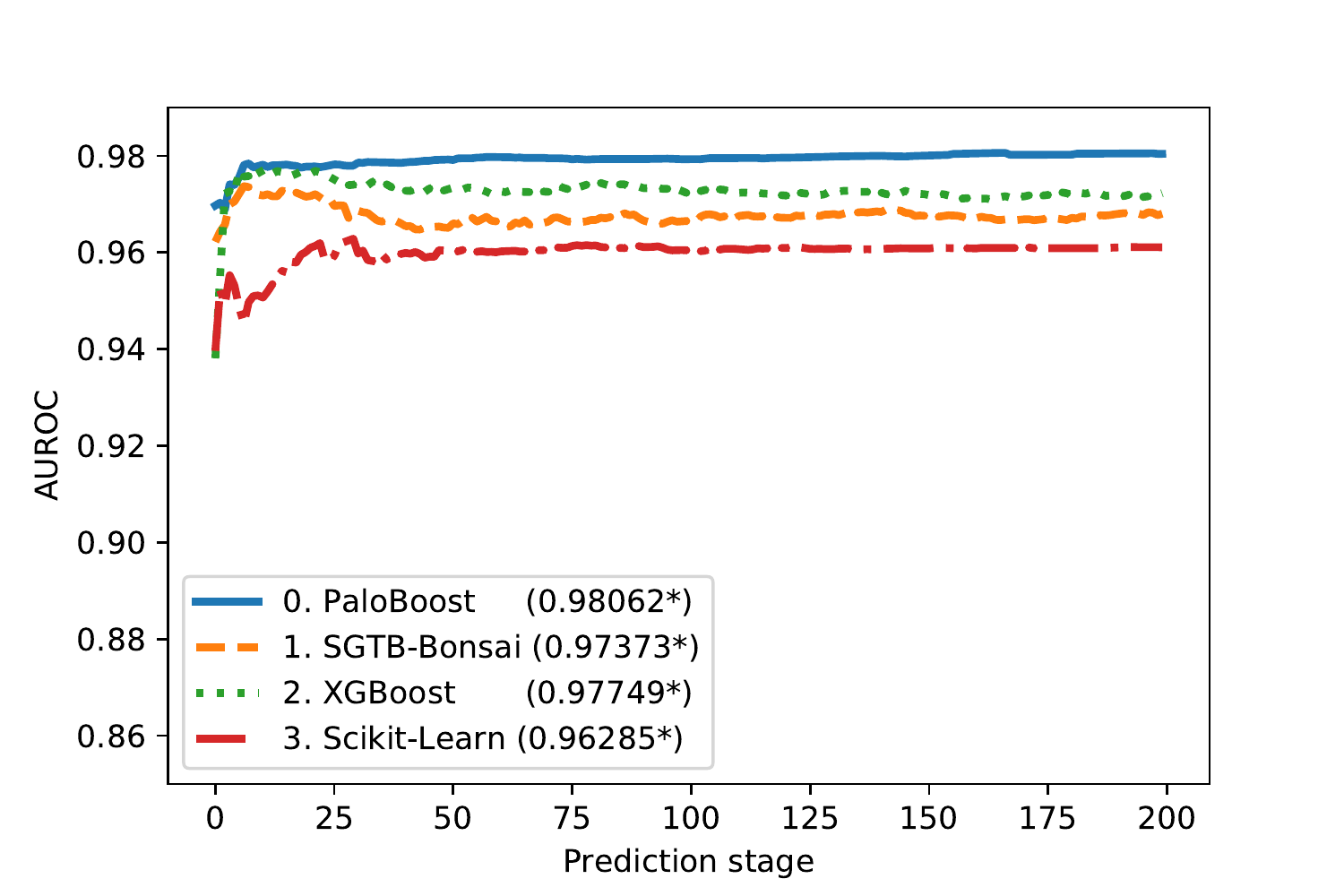}}
        \subfloat[Pulsar $\nu_{max}=0.1$]{\includegraphics[width=0.33\textwidth]{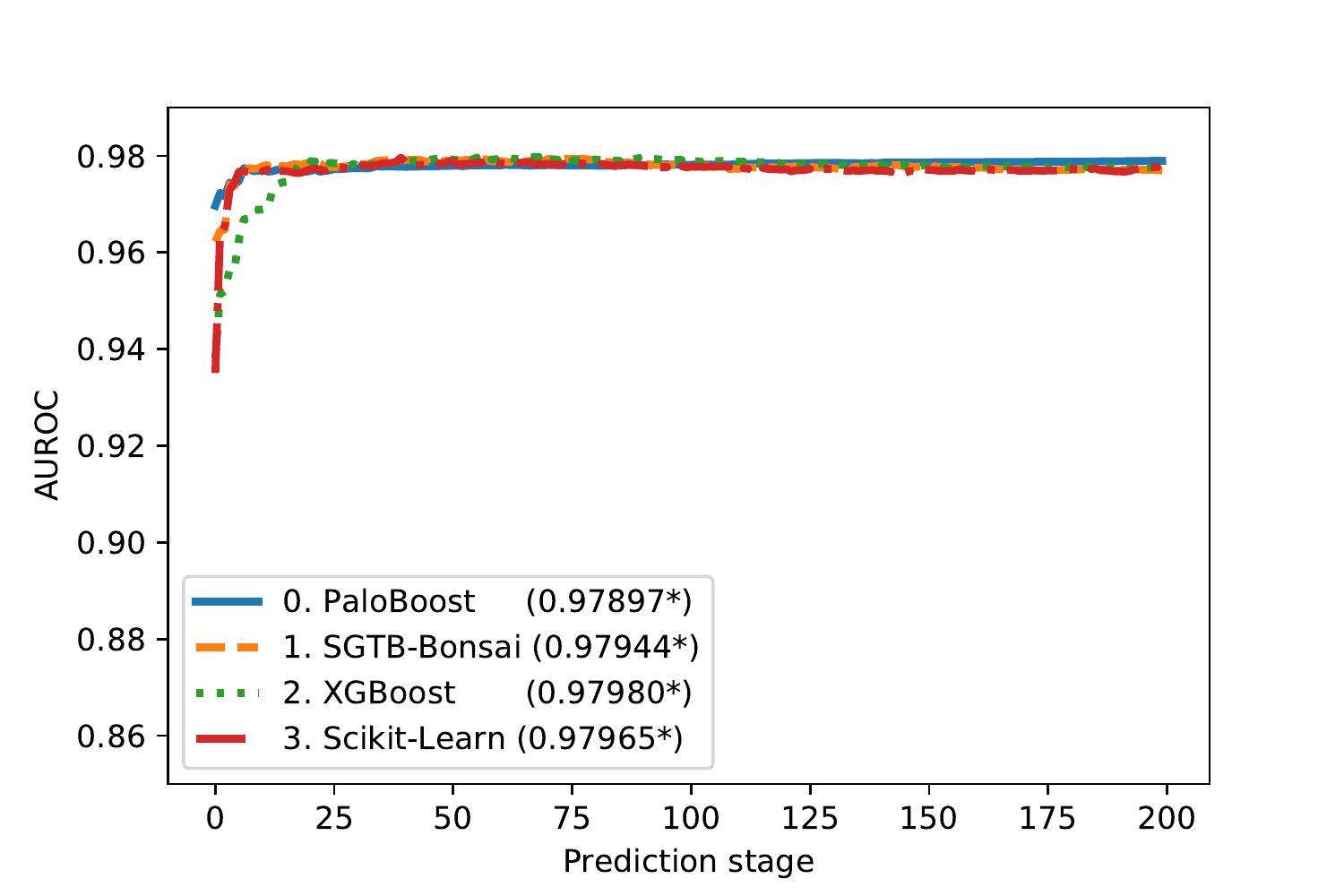}} \\
        \subfloat[Carvana, $\nu_{max}=1.0$]{\includegraphics[width=0.33\textwidth]{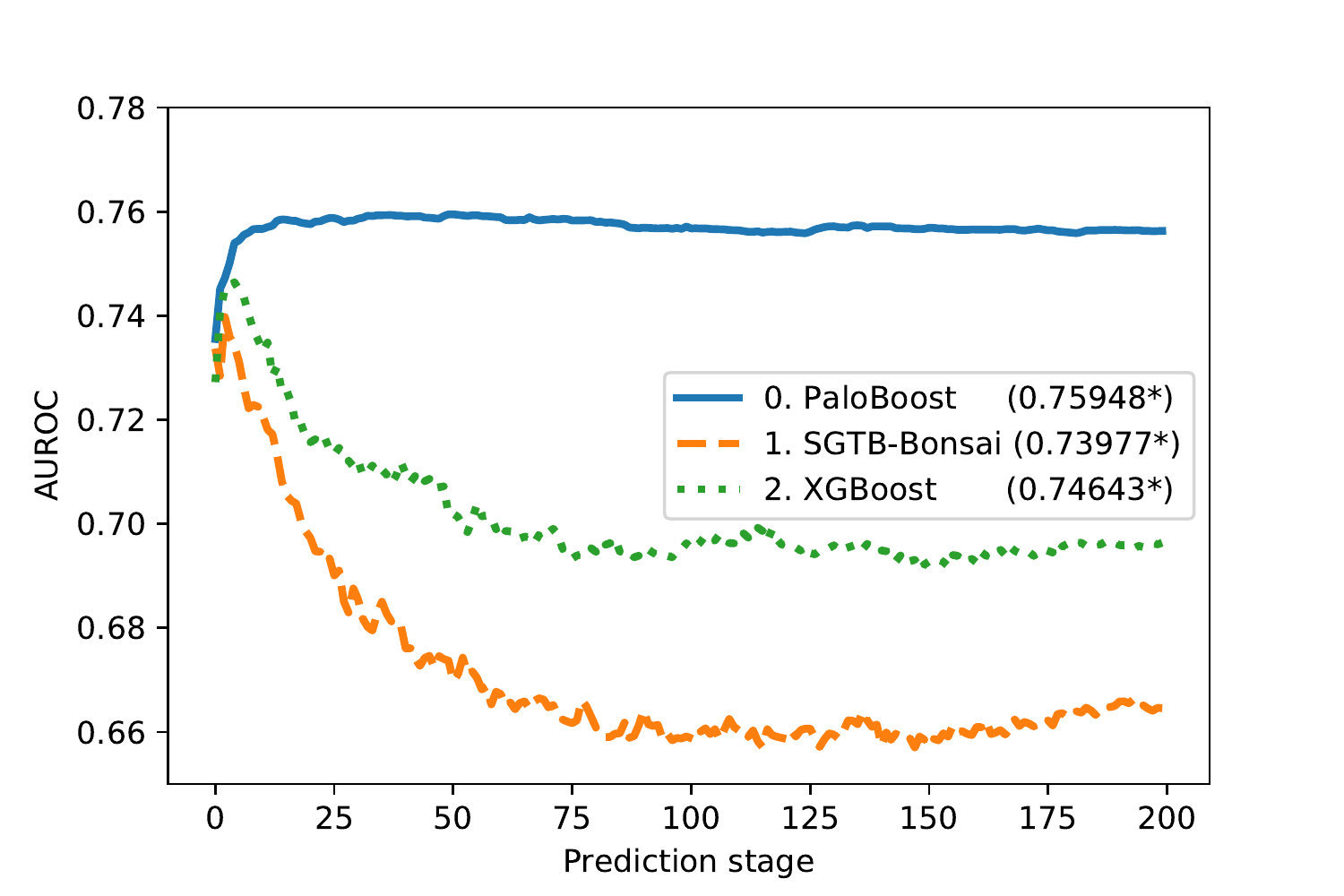}}
        \subfloat[Carvana, $\nu_{max}=0.5$]{\includegraphics[width=0.33\textwidth]{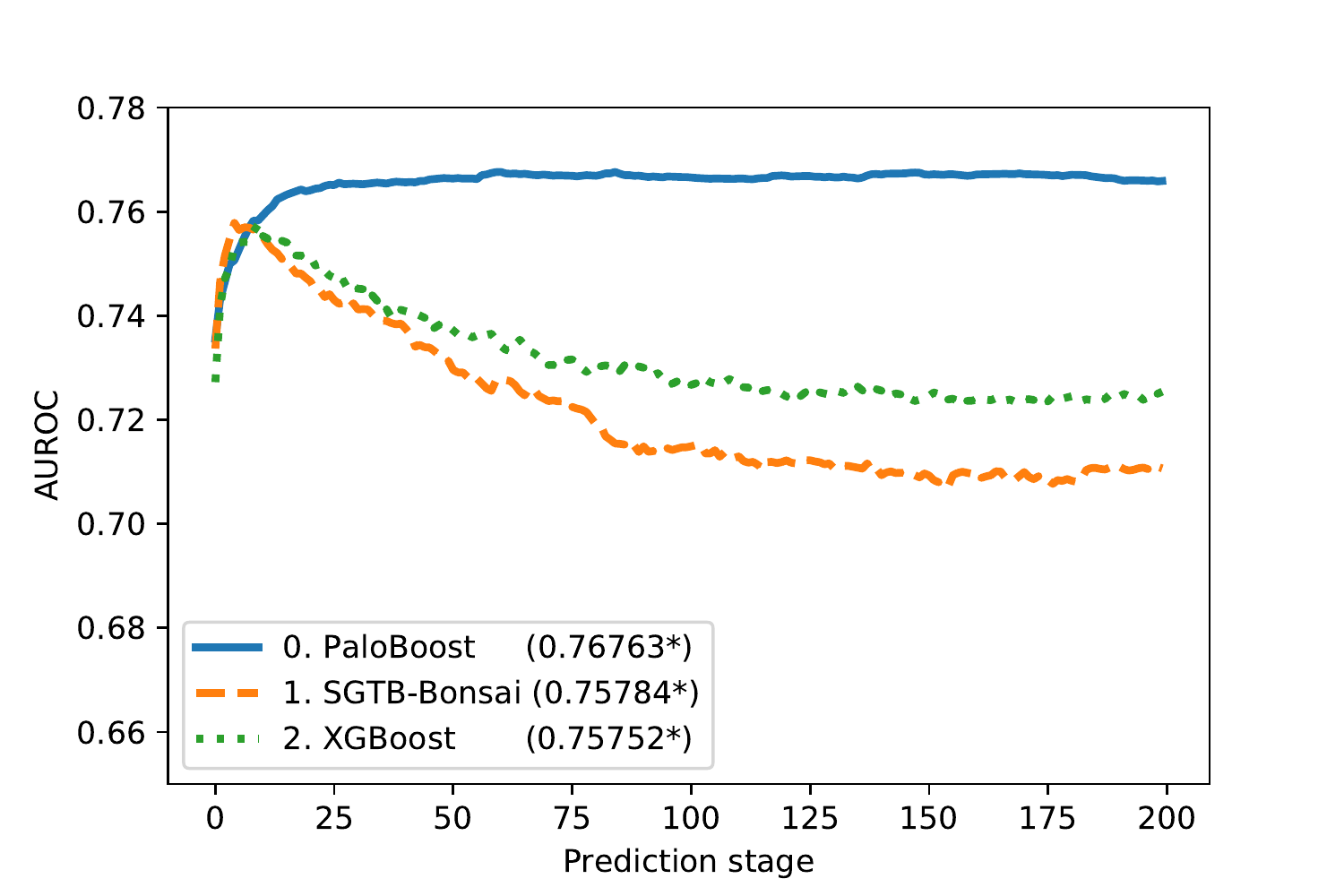}}
        \subfloat[Carvana, $\nu_{max}=0.1$]{\includegraphics[width=0.33\textwidth]{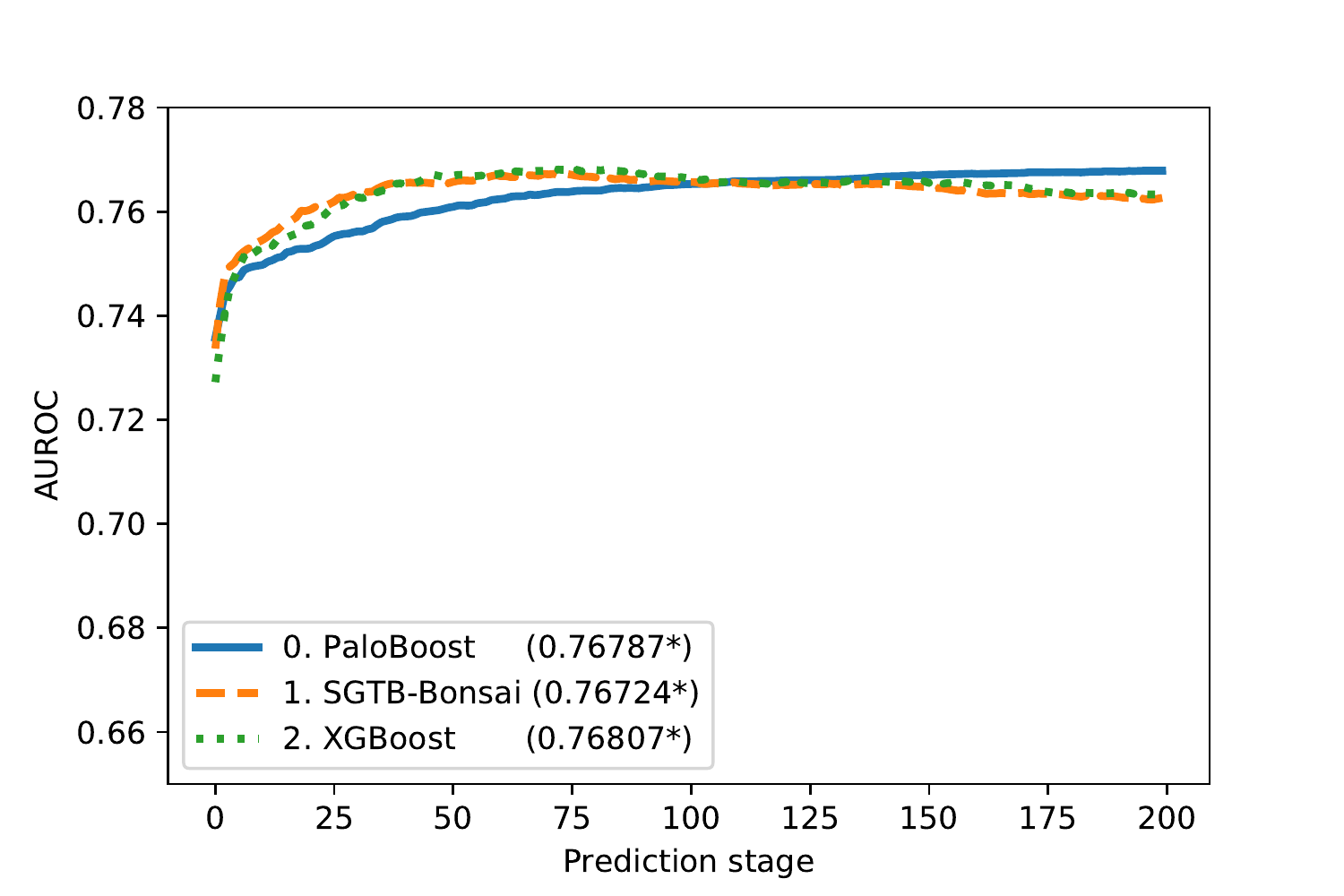}} \\
        \subfloat[BNP, $\nu_{max}=1.0$]{\includegraphics[width=0.33\textwidth]{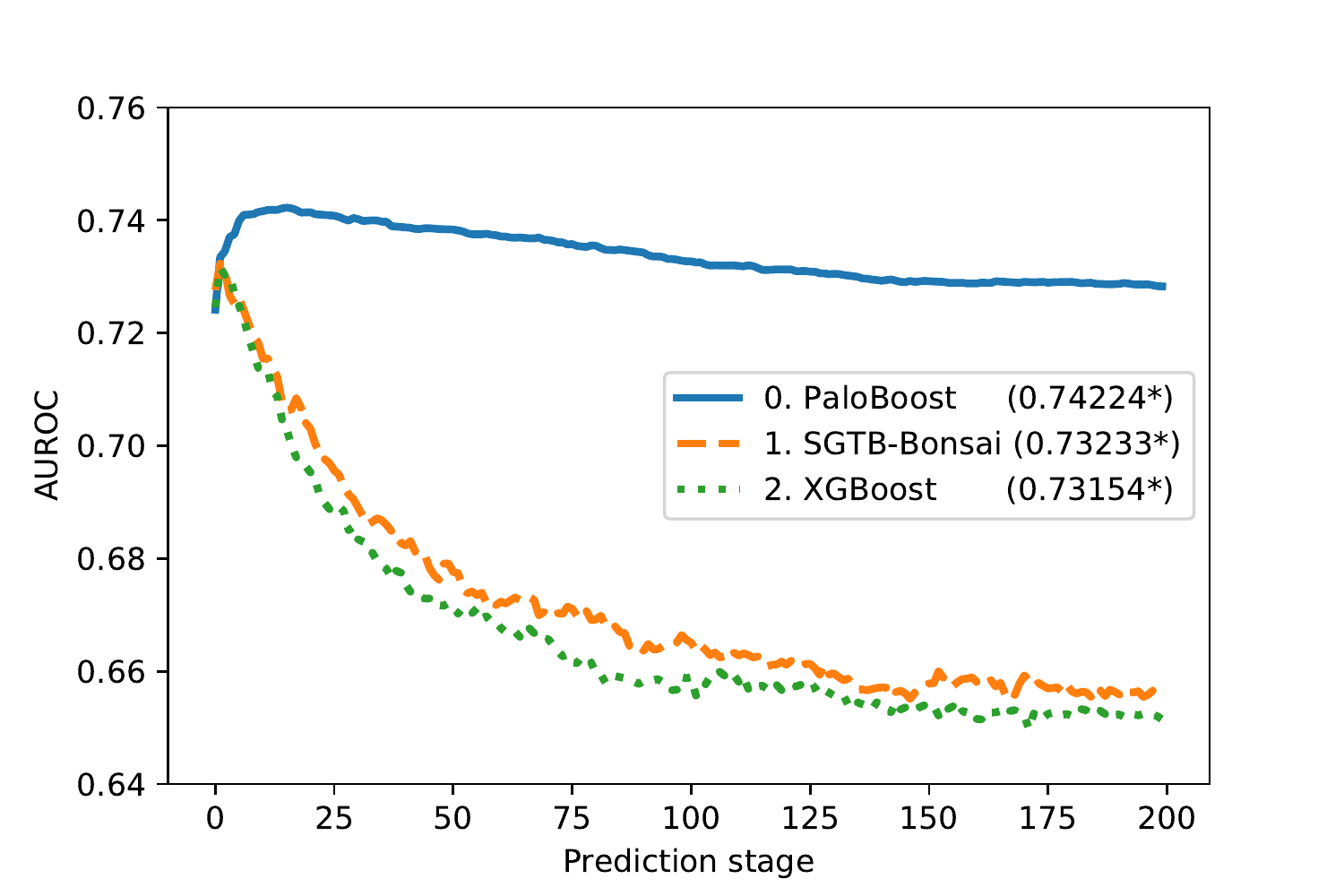}}
        \subfloat[BNP, $\nu_{max}=0.5$]{\includegraphics[width=0.33\textwidth]{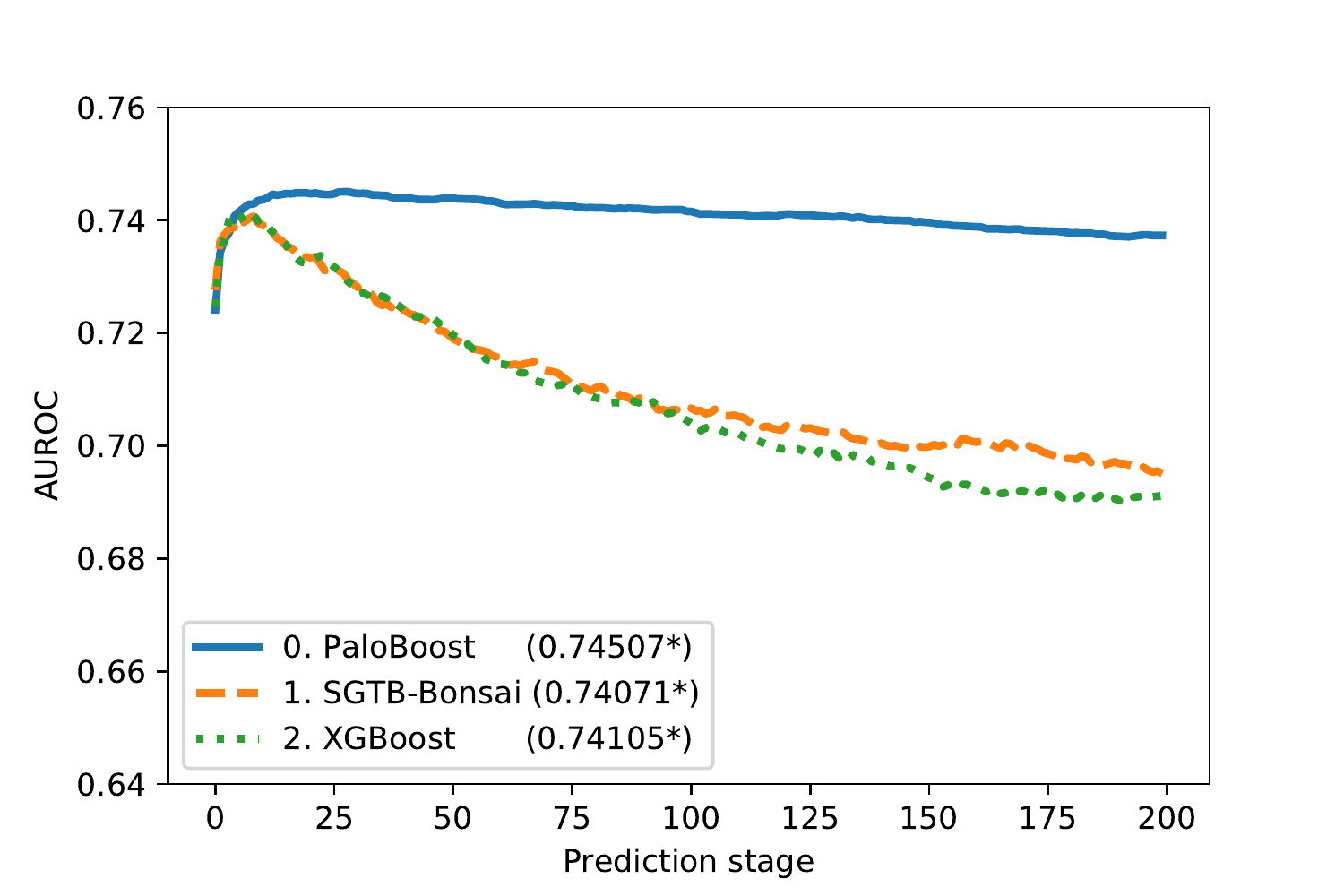}}
        \subfloat[BNP, $\nu_{max}=0.1$]{\includegraphics[width=0.33\textwidth]{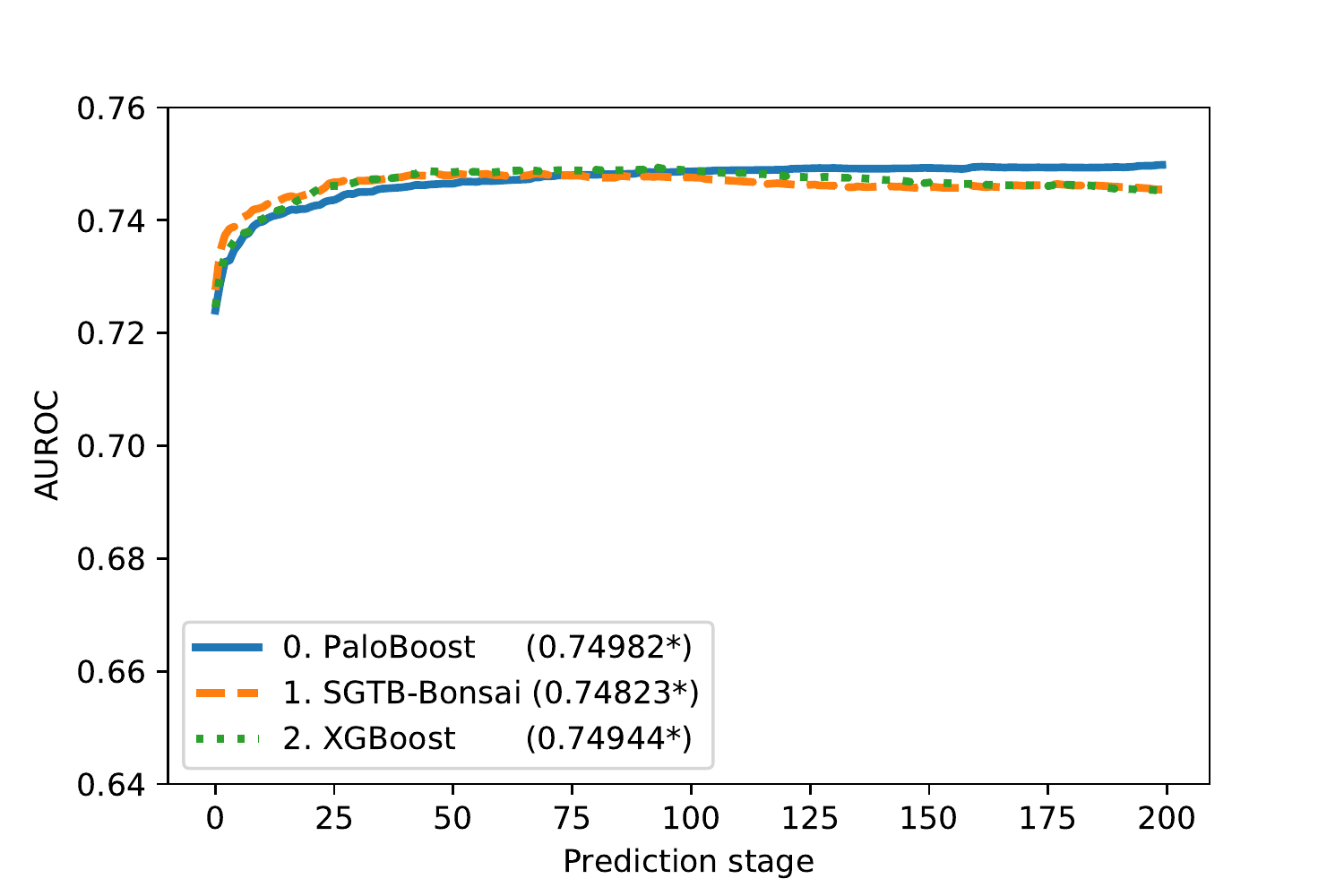}}
        \caption{
        Predictive performance, measured in AUROC, for the four classification tasks.
        From the top, at each row, we have Amazon Employee Access Dataset (1st row),  Pulsar Detection Dataset (2nd row), Carvana Dataset (3rd row), and BNP-Paribas Cardif Dataset (4th row).
        The y-axis is aligned for each dataset (row) for ease of comparison across learning rates.
        }
    \label{fig:classification}
\end{figure*}

\subsubsection{Amazon Employee Access Dataset}
Depending on their roles, Amazon's employees have different access rights on their internal web resources.
Supervisors often spend a considerable amount of time manually granting and revoking the access rights of their employees.
To automate this process and increase the overall work efficiency,
Amazon released its historical employee access dataset on Kaggle\footnote{https://www.kaggle.com/c/amazon-employee-access-challenge} that contains employee attributes, resource ID, and whether the access was granted or not (target).
Interestingly, the dataset only contains eight categorical features (preprocessed to 115 features) for 32,769 employees (see Table \ref{tab:data}).
Also, many top-scoring solutions on the Kaggle leaderboard noted the importance of capturing the interaction between categorical features to improve the predictive performance.
This is due to the fact that the employees' access rights are somewhat well defined given a specific configuration of employee attributes.
Therefore, there exists a fairly deterministic relationship between the features and the target, which can be obtained by a highly customized preprocessing step.
However, the performance improvement is not the key objective of our experiments, so minimal preprocessing is performed.

The topmost row in Figure~\ref{fig:classification} presents the results from the Amazon Employee Dataset experiment.
Unlike the previous datasets (simulated, Mercedes-Benz, and Crime), the AUROC of PaloBoost, SGTB-Bonsai, and XGBoost are still improving after 200 iterations.
Only Scikit-Learn demonstrate overfitting on the dataset.
We suspect there are two reasons for this behavior.
Since all the features in this dataset are categorical, the degree of freedom in the input space is finite, thereby reducing the risk of the curse of dimensionality.
Secondly, the fairly deterministic nature of the features and the targets combined with the large sample size make it less prone to overfitting.

Another important observation from Figure \ref{fig:classification} is that PaloBoost converges slower compared to the other models on this dataset.
This is because PaloBoost has two additional regularization mechanisms: gradient-aware pruning and adaptive learning rates.
While these modifications are likely to improve the performance for a complex problem (e.g., large number of features, small dataset size, large amount of noise, etc.), it can hinder the performance on simpler problems.
By regularizing the trees and the learning rates, PaloBoost is unable to quickly learn an appropriate model and thereby requires more iterations to achieve the same accuracy\footnote{PaloBoost eventually achieves comparable AUROC.}.
Therefore, PaloBoost may not be appropriate when standard SGTB models are not prone to overfitting.

\subsubsection{Pulsar Detection Dataset}
Pulsars, rapidly rotating neutron stars, are known to emit a detectable pattern of electromagnetic radiation.
Unfortunately, automated detection based on the pattern is often a non-trivial task due to radio frequency interference and noise.
Nowadays, machine learning techniques are adopted to extract candidates for detecting pulsars.
In the effort to develop such machine learning detection algorithms, 
Lyon curated and published the HTRU2 (High Time Resolution Universe Survey - South) dataset \cite{Lyon:2016, Lyon:data}.
The dataset, published on UCI\footnote{https://archive.ics.uci.edu/ml/datasets/HTRU2}, contains the profiles of electromagnetic waves and the label (pulsar or not) of 17,898 examples, 
where 1,639 examples are pulsars (positive class). 
All eight features are numeric and have no missing values (see Table \ref{tab:data}).

The second row in Figure~\ref{fig:classification} illustrates the results from the Pulsar Detection Dataset experiment.
As can be seen, PaloBoost exhibits the most stable and best performance overall.
However, the performance differences are practically indistinguishable for smaller learning rates.
This may be a result of the dataset curation process, where only engineered features that are strongly correlated with the target variable were included.
Even a single, albeit highly customized, decision tree algorithm could achieve a reasonable performance \cite{Lyon:2016}. 
In other words, the dataset has no noisy features, thus overfitting is less likely to happen even for complex models.

\subsubsection{Carvana Dataset}

When auto dealers purchase used cars from an auction, 
there is a risk of purchasing cars with serious issues such as tampered odometers and mechanical issues, 
which is commonly referred as ``kicks`` in the industry.
Carvana is an online used-car dealers company. 
The company released its dataset on Kaggle\footnote{https://www.kaggle.com/c/DontGetKicked} that contains used cars' attributes, such as make, model, vehicle age, odometer, etc., and the target variable (kick or not). 
The dataset (summarized in Table \ref{tab:data}) has 72,983 samples, with 151 numeric features (after preprocessing).
Due to missing values, Scikit-Learn is omitted from this experimental study.

The third row in Figure~\ref{fig:classification} shows the results from the dataset.
As can be seen, the two baseline models clearly display overfitting behaviors for all three learning rates.
Moreover, the dataset is prone to overfitting, as illustrated by the Kaggle leaderboards.
A comparison of the public and private leaderboards\footnote{Kaggle sets one-third of test data for ranking the public leaderboard, and the rest for the private leaderboard. The private leaderboard is closed during the competition and announced after the competition is closed. This is to prevent the overfitting on the test set \cite{Blum:2015}.} shows considerable movement between the top 4 teams. 
This indicates the top teams' models likely suffered from overfitting.
Notably, PaloBoost exhibits extremely stable predictive performances and records the best scores for all cases.
Moreover, the difference in performance across the three learning rates is almost negligible.
Although not shown, the prune and adaptive learning rates displayed similar patterns to the simulated dataset (Figure \ref{fig:friedman-palo}).

\subsubsection{BNP Paribas Cardif Claims Dataset}

Insurance claim payments involve many levels of manual checks, sometimes requires more information to be collected, 
and can take a substantial amount of time till the final payments.
Some of these claims, however, can be processed much faster with machine learning techniques.
BNP Paribas Cardif, one of the largest personal insurance company, released some of its data via the Kaggle platform\footnote{https://www.kaggle.com/c/bnp-paribas-cardif-claims-management}
that contains various claim attributes (features) and if the claim's approval could be accelerated or not (target).
The dataset consists of 114,321 samples with 273 numeric features (after preprocessing).
Scikit-Learn is omitted from this study as the dataset has many missing values (see Table \ref{tab:data}).
This dataset contains the largest number of samples (of the seven datasets) and shows a fair degree of complexity. 

The last row in Figure~\ref{fig:classification} illustrates the results from the BNP Paribas Cardif Claims Dataset experiment.
Similar to all but the Amazon classification task, PaloBoost offers the most stable and best predictive performance.
The competitors involved in this Kaggle challenge remarked that they suspect the dataset has additional noise to the features\footnote{See discussions 19240 and 20247 on the Kaggle discussion board at \url{https://www.kaggle.com/c/bnp-paribas-cardif-claims-management/discussion/} as some examples.}.
Many winning solutions were based on extensive feature engineerings, identifying the relationships of the features, and removing the noisy features.
The impressive performance from PaloBoost suggests the ability to effectively identify the noisy features to yield a more robust model.
This also suggests that PaloBoost can outperform other SGTB implementations when a minimal preprocessing is applied to a noisy dataset.

Based on our extensive experiments, PaloBoost outperforms existing SGTB models when a dataset has many noisy features and is prone to overfitting.
While PaloBoost is not guaranteed to achieve the best predictive performance on all datasets (see the Amazon results in the top row of Figure \ref{fig:classification}), it is less prone to overfitting.
From Figures \ref{fig:friedman-perf}, \ref{fig:regression}, and \ref{fig:classification}), we observe that PaloBoost exhibits a gradual performance degradation in all seven datasets, unlike the other baseline models.
Moreover, the predictive performance of PaloBoost using different learning rates is relatively small.
Thus, we conclude that PaloBoost is robust to the specification of the learning rate ($\nu_{max}$), the tree depth, and also the number of iterations $M$.

\section{Discussions}

We introduced PaloBoost, an extension of SGTB, to mitigate overfitting and minimize exhaustive hyperparameter tuning.
PaloBoost uses two regularization techniques to perform gradient-aware pruning and adaptive learning rate estimation.
Rather than viewing OOB samples as a third-party observer for tracking errors and feature importances, PaloBoost considers these samples as an alternative training sample.
Based on this new perspective of the under-utilized OOB samples, PaloBoost can dynamically adjust the tree depths and learning rates at each stage.
With these two mechanisms, PaloBoost can automatically adapt to minimize overfitting by knowing when it needs to ``slow down".
Furthermore, by introducing efficient computations, these regularizations can be readily implemented on top of SGTB with minimal computational impact.

Our extensive experiments using both real and simulated datasets confirm that PaloBoost produces robust predictive performance.
In particular, when a dataset is noisy and complex, PaloBoost significantly outperforms the other SGTB implementations.
We also show that our two regularization mechanisms can adaptively adjust to guard against overfitting by yielding lower variance trees, and optimal region-specific learning rates. 
Moreover, the empirical results demonstrate considerably less sensitivity to the parameter settings -- different learning rates and the number of iterations yield comparable predictive performance.

We also introduced a new feature importance formula that showed promising results on the Friedman's Simulated Dataset.
While other SGTB implementations failed to identify the relevant features, PaloBoost's importance estimates accurately captured the true importances.
We posit that the OOB regularizations (removing noisy nodes) in conjunction with accounting for the coverage of the region and the node estimates, yields a superior feature selection process.
In addition, we proposed a new visualization based on feature importance to identify the optimal number of iterations.
It would be interesting to see if the feature selection mechanisms from PaloBoost can help other machine learning algorithms.

Another observation drawn from the experiments was the presence of many trees with negligible (close to zero) learning rates in PaloBoost.
Given that these trees have minimal impact on the overall performance, they can be potentially removed.
This should yield a more compact-sized tree boosting model while offering similar predictive performance.
Perhaps this can be further extended to encompass a two-steps-forward-one-step-back strategy.
By integrating tree removal directly into the algorithm, a more cohesive and compact model can be learned without introducing significant computational overhead.
These intriguing ideas are left for future work.

PaloBoost is developed to make the SGTB more robust and stable. 
While significantly less sensitive to the hyperparameters compared to the other implementations, PaloBoost still requires similar parameters (``max'' learning rate and tree depth).
We note that the ideas presented in this paper are the initial steps towards automating the tuning of SGTB models.
Not only can PaloBoost save computation resources and researchers' time, but it can also help democratize SGTB to a wider audience.

\bibliographystyle{ACM-Reference-Format}
\bibliography{paloboost}

\end{document}